%% file: icml_main.tex
\documentclass[11pt]{article}
\usepackage{verbatim}
\usepackage{amsmath,amsbsy,amsfonts,amssymb,amsthm,dsfont,fullpage,units}
\usepackage{graphicx}
\usepackage{algorithm,algorithmic,mathtools}
\usepackage{color}
\usepackage{tikz}
\usepackage{float}
\usetikzlibrary{calc,shapes}
\usepackage{natbib}
\usepackage{caption}
\usepackage{subcaption}
\usepackage{hyperref}

\newtheorem{claim}{Claim}
\newtheorem{corollary}{Corollary}[section]

\newtheorem{lemma}{Lemma}
\newtheorem{proposition}{Proposition}[section]
\newtheorem{theorem}{Theorem}[section]
\theoremstyle{definition}
\newtheorem{condition}{Condition}[section]
\newtheorem{definition}{Definition}

\newtheorem{remark}{Remark}

 \newcommand{\IGNORE}[1]{}

\newcommand\E{\mathbb{E}}
\newcommand\Var{\mbox{Var}}
\newcommand\R{\mathbb{R}}
\newcommand\N{\mathcal{N}}
\newcommand\calD{\mathcal{D}}

\newcommand\eps{\varepsilon}

\newcommand\rcu{\hat{\mu}}
\newcommand\icu{\mu_{i\rightarrow}}

\newcommand\inner[1]{\langle #1 \rangle}

\title{Stronger generalization bounds for deep nets via a compression approach}
\date{}
\author{Sanjeev Arora\thanks{Princeton University, Computer Science Department, email: arora@cs.princeton.edu} \and Rong Ge\thanks{Duke University, Computer Science Department, email: rongge@cs.duke.edu}\and Behnam Neyshabur\thanks{Institute for Advanced Study, School of Mathematics, email: bneyshabur@ias.edu}\and Yi Zhang\thanks{Princeton University, Computer Science Department, email: y.zhang@cs.princeton.edu}}

\begin{document}
\maketitle
\begin{abstract}
Deep nets generalize well despite having more parameters than the number of training samples. Recent works try to give an explanation using PAC-Bayes and Margin-based analyses, but do not as yet result in sample complexity bounds better than naive parameter counting. 
The current paper shows generalization bounds that're orders of magnitude better in practice. These rely upon new succinct reparametrizations of the trained net --- a compression that is explicit and efficient. These yield generalization bounds via a simple compression-based framework introduced here. Our results also provide some theoretical justification for widespread empirical success in compressing deep nets.

 Analysis of correctness of our compression relies upon some newly identified \textquotedblleft noise stability\textquotedblright   properties of trained deep nets, which are also experimentally verified. 
The study of these properties and resulting generalization bounds are also extended to convolutional nets, which had eluded  earlier attempts on proving generalization. 

\end{abstract}

\input{intro}

\input{compression}

\input{margin}

\input{newmatrix}
\input{convolutionmain}
\input{experiments}

\input{conclusions}

\section*{Acknowledgements}
This research was done with support from NSF, ONR, Simons Foundation, Mozilla Research, and Schmidt Foundation.
\bibliographystyle{plainnat}
\bibliography{ref}
\clearpage
\onecolumn
\input{appendix}

\input{fullyconnected}
\input{convolution}

\input{extended_experiments}

\end{document}

%% file: intro.tex
\section{Introduction}
A mystery about deep nets is that they  generalize (i.e., predict well on unseen data) despite having far more parameters than the number of training samples. One commonly voiced explanation is that regularization during training \---whether implicit via use of SGD~\cite{neyshabur15b,hardt2015train} or explicit via weight decay, dropout~\cite{srivastava2014dropout}, batch normalization~\cite{IofSze15}, etc. \---reduces the effective capacity of the net. But \citet{zhang2017understanding} questioned this received wisdom and fueled research in this area by showing experimentally that standard architectures using SGD and regularization can still reach low training error on randomly labeled examples (which clearly won't generalize).

Clearly, deep nets trained on real-life data have some properties that reduce effective capacity, but identifying them has proved difficult ---at least in a {\em quantitative} way that yields sample size upper bounds similar to classical analyses in simpler models such as SVMs~\cite{bartlett2002rademacher,evgeniou2000regularization,smola1998connection} or matrix factorization ~\cite{fazel2001rank,srebro2005maximum}. 

Qualitatively~\cite{hochreiter1997flat,hinton1993keeping} suggested that  nets that generalize well are {\em flat minima} in the optimization landscape of the training loss and \cite{hinton1993keeping} further discusses the connections to Minimum Description Length principle for generalization. Recently \citet{keskar2016large}  show using experiments with different batch-sizes that sharp minima do correlate with higher generalization error. A quantitative version of \textquotedblleft flatness\textquotedblright\ was suggested in~\citep{langford2001not}: the net's output is stable to {\em noise} added to the net's trainable parameters. Using PAC-Bayes bound~\citep{mcallester1998some,mcallester1999pac} this noise stability yielded generalization bounds for fully connected nets of depth $2$.  The theory has been extended to multilayer fully connected nets~\citep{neyshabur2017exploring}, although thus far yields sample complexity bounds much worse than naive parameter counting. (Same holds for the earlier~\cite{bartlett2002rademacher,NeyTomSre15,bartlett2017spectrally,neyshabur2017pac,golowich2017size}; see Figure~\ref{fig:comparison}). Another notion of noise stability ---closely related to dropout and batch normalization---is stability of the output with respect to the noise injected at the nodes of the network, which was recently shown experimentally~\citep{morcos2018} to improve in tandem with generalization ability during training, and to be absent in nets trained on random data. Chaudhari et al~\cite{chaudhari2016entropy} suggest adding noise to gradient descent to bias it towards finding flat minima.

 While study of generalization may appear a bit academic --- held-out data easily establishes generalization in practice--- the  ultimate hope is that it will help identify simple, measurable and intuitive properties of well-trained deep nets, which in turn may fuel  superior architectures and faster training. We hope the detailed study ---theoretical and empirical---in the current paper advances this goal.

\begin{figure}[ht]
	\centering
	\includegraphics[width=16cm]{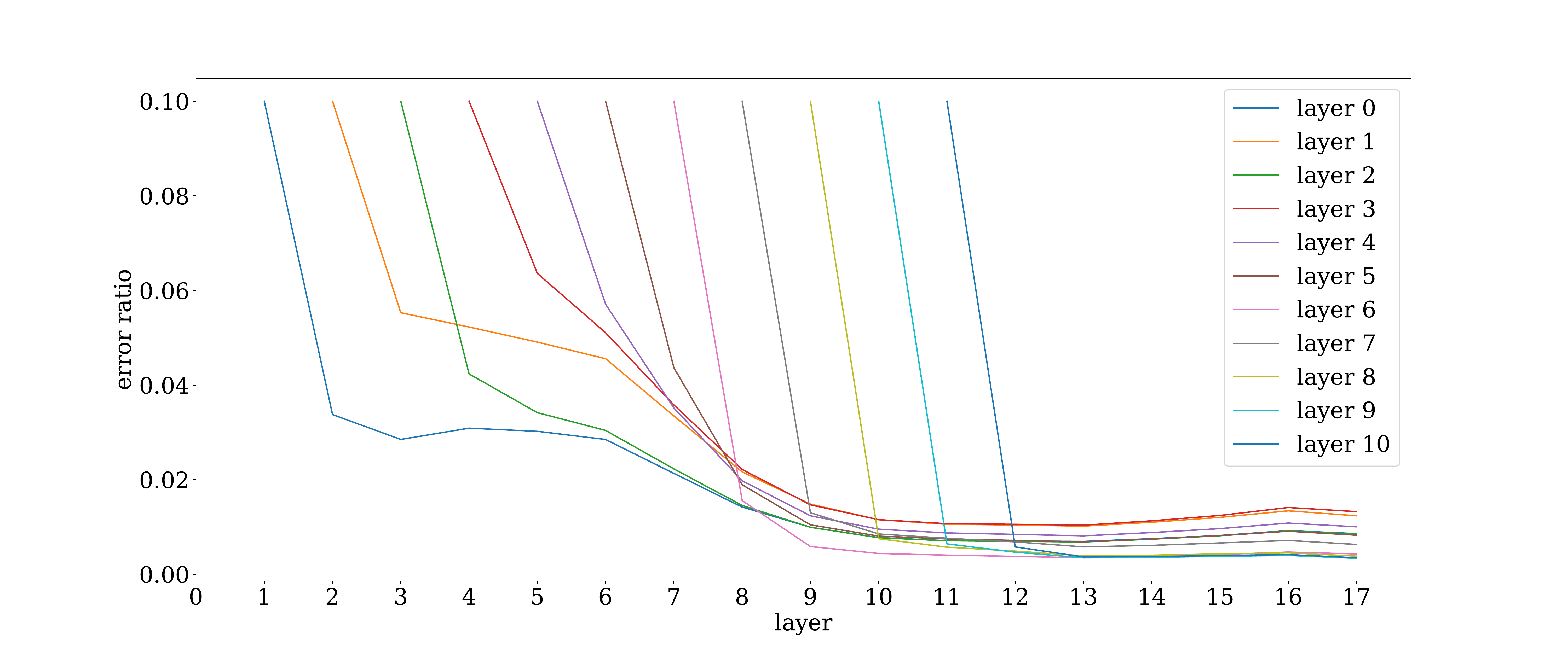}
	\caption{Attenuation of injected noise on a VGG-19 net trained on CIFAR-10. The x-axis is the index of layers and y-axis denote the relative error due to the noise ($\|\hat{x}^i - x^i\|_2/\|x^i\|_2$). A curve starts at the layer where a scaled Gaussian noise is injected to its input, whose $\ell_2$ norm is set to $10\%$ of the norm of its original input. As it propagates up, the injected noise has rapidly decreasing effect on higher layers. This property is shown to imply compressibility. }
	\label{fig:noise_decay}
	\vspace{-0.5cm}
\end{figure}

 \noindent{\bf Contributions of this paper.} 
 \vspace*{-0.1in}
 \begin{enumerate}
 \item A simple {\em compression framework} (Section~\ref{sec:compress}) for proving generalization bounds, perhaps a more explicit and intuitive form of the PAC-Bayes work. It also yields elementary short proofs of recent generalization results~(Section~\ref{subsec:existing}).
 \item Identifying new form of noise-stability for deep nets: the stability of each layer's computation to noise injected at lower layers. (Earlier papers worked only with stability of the {\em output} layer.) Figure~\ref{fig:noise_decay} visualizes the stability of network w.r.t. Gaussian injected noise. Formal statements require a string of other properties (Section~\ref{sec:stability}). All are empirically studied, including their correlation with generalization (Section~\ref{sec:experiments}). 
 \item Using the above properties to derive efficient and {\em provably correct} algorithms that reduce the effective number of parameters in the nets, yielding generalization bounds that: (a) are better than naive parameter counting (Section~\ref{sec:experiments}) (b) depend on simple, intuitive and measurable properties of the network (Section~\ref{sec:newmatrix}) (c) apply also to convolutional nets (Section~\ref{sec:convolution}) (d) empirically correlate with generalization (Section~\ref{sec:experiments}).
 \end{enumerate}

The main idea is to show that noise stability allows individual layers to be compressed via a linear-algebraic procedure Algorithm~\ref{alg:matrix-proj}. This results in new error in the output of the layer. This added error is \textquotedblleft gaussian-like\textquotedblright\ and tends to get attenuated as it propagates to higher layers.

\medskip

\noindent{\bf Other related works.}
\citet{dziugaite2017computing}  use non-convex optimization to optimize the PAC-Bayes bound  get a non-vacuous sample bound on MNIST. 
While very creative, this  provides little insight into favorable properties of networks. \citet{liang2017fisher} have suggested Fisher-Rao metric, a regularization based on the Fisher matrix and showed that this metric correlate with generalization.  Unfortunately, they could only apply their method to linear networks. Recently \citet{kawaguchi2017generalization} connects Path-Norm \cite{neyshabur2015path} to generalization. However, the proved generalization bound depends on the distribution and measuring it requires vector operations on exponentially high dimensional vectors. 
Other works have designed experiments to empirically evaluate potential properties of the network that helps generalization\cite{arpit2017closer,neyshabur2017exploring,dinh2017sharp}.
The idea of compressing trained deep nets is very popular for low-power applications; for a survey see~\citep{compresssurvey}.

Finally, note that the terms {\em compression} and {\em stability} are traditionally used in a different sense in generalization theory~\cite{littlestone1986relating,kearns1999algorithmic,shalev2010learnability}. Our framework  is compared to other notions in the remarks after Theorem~\ref{thm:compress_generalize}.


\medskip

\noindent{\bf Notation:} We use standard formalization of  multiclass classification, when data has to be assigned a {\em label} $y$ of a sample $x$ which is an integer from $1$ to $k$. A multiclass classifier $f$ maps input $x$ to $f(x) \in \R^k$ and the classification loss for any distribution $\calD$ is defined as
$\mathbb{P}_{(x,y)\sim \calD} \left[f(x)[y] \leq  \max_{i\neq y} f(x)[j]\right].$
If  $\gamma>0$ is some desired margin, then the expected margin loss is
$$
L_{\gamma}(f) = \mathbb{P}_{(x,y)\sim \calD} \left[f(x)[y] \leq \gamma + \max_{i\neq y} f(x)[j]\right]
$$
(Notice, the classification loss corresponds to $\gamma=0$.)  Let $\hat{L}_{\gamma}$ denote empirical estimate of the  margin loss. {\em Generalization error} is the difference between the two.

For most of the paper we assume that deep nets have fully connected layers, and use ReLU activations. We treat convolutional nets in Section~\ref{sec:convolution}. If the net has $d$ layers, we label the vector before activation at these layers by $x^0$, $x^1$, … $x^d$ for the $d$ layers where $x^0$ is the input to the net, also denoted simply $x$. 
So  $x^{i} = A^{i} \phi(x^{i-1})$ where $A^i$ is the weight matrix of the $i$th layer. (Here $\phi(x)$ if $x$ is a vector applies the ReLU component-wise. The ReLU is allowed a trainable bias parameter, which is omitted from the notation because it has no effect on any calculations below.) We denote the number of hidden units in layer $i$ by $h^i$ and set $h=\max_{i=1}^d h^i$. Let $f_A(x)$ be the function calculated by the above network.

{\em Stable rank} of a matrix $B$ is $\|B\|_F^2/\|B\|_2^2$, where $\|\cdot\|_F$ denotes Frobenius norm and 
$\|\cdot \|_2$ denotes spectral norm. Note that stable rank is at most (linear algebraic) rank.

For any two layer $i\leq j$, denote by $M^{i,j}$ the operator for composition of these layers and $J^{i,j}_x$ be the Jacobian of this operator at input $x$ (a matrix whose $p, q$ is the partial derivative of the $p$th output coordinate with respect to the $q$'th input input). Therefore, we have $x^j = M^{i,j}(x^i)$. Furthermore, since the activation functions are ReLU, we have $M^{i,j}(x^i) =J^{i,j}_{x^i} x^i$.

%% file: compression.tex
\section{Compression and Generalization}
\label{sec:compress}


Our compression framework rests on the following obvious fact. Suppose the training data contains $m$ samples, and $f$ is a classifier from a complicated class (e.g., deep nets with much more than $m$ parameters) that incurs very low empirical loss. We are trying to understand if it will generalize. Now suppose we can compute a classifier $g$ with discrete trainable parameters much less than $m$ and which incurs  similar loss on the training data as $f$.  Then $g$ must incur low classification error on the full distribution. 
This framework has the advantage of staying with intuitive parameter counting and to avoid explicitly dealing with the hypothesis class that includes $f$ (see note after Theorem~\ref{thm:compress_generalize}).
Notice, the mapping from $f$ to $g$ merely needs to {\em exist,} not to be efficiently computable. But in all our examples the mapping will be explicit and fairly efficient. Now we formalize the notions. The proofs are elementary via concentration bounds and appear in the appendix.

\begin{definition}[($\gamma$,$S$)-compressible]\label{def:compressible} Let $f$ be a classifier and $G_{\mathcal{A}}=\{g_A|A\in\mathcal{A}\}$ be a class of classifiers. We say $f$ is ($\gamma,S$)-compressible via $G_{\mathcal{A}}$ if there exists $A\in \mathcal{A}$ such that for any $x\in S$, we have for all $y$
$$
|f(x)[y] - g_A(x)[y]| \le \gamma.
$$
\end{definition}

We also consider a different setting  where the compression algorithm is allowed a``helper string'' $s$, which is arbitrary but fixed before looking at the training samples. Often  $s$ will contain random numbers. A simple example is to let $s$ be the random initialization used for training the deep net. Then compress the {\em difference} between the final weights and $s$; this can give better generalization bounds, similar to~\cite{dziugaite2017computing}. Other nontrivial examples appear later.


\begin{definition}[($\gamma$,$S$)-compressible using helper string $s$] Suppose $G_{\mathcal{A},s} =\{g_{A,s}|A\in \mathcal{A}\}$ is a class of classifiers indexed by trainable parameters $A$ and fixed strings $s$.  A classifier $f$ is  ($\gamma,S$)-compressible with respect to $G_{\mathcal{A},s}$ using helper string $s$ if there exists $A\in \mathcal{A}$ such that for any $x\in S$, we have for all $y$
\begin{equation*}
	|f(x)[y] - g_{A,s}(x)[y]| \le \gamma.
\end{equation*}
\end{definition}

\begin{theorem} \label{thm:compress_generalize} Suppose $G_{\mathcal{A},s} =\{g_{A,s}|A\in \mathcal{A}\}$ where $A$ is a set of $q$ parameters each of which can have at most $r$ discrete values and $s$ is a helper string. Let $S$ be a training set with $m$ samples. If the trained classifier $f$ is $(\gamma,S)$-compressible via $G_{\mathcal{A},s}$ with helper string $s$, then there exists $A\in\mathcal{A}$ with high probability over the training set,
	$$
	L_0(g_A) \le \hat{L}_\gamma(f)+ O\left(\sqrt{\frac{q\log r}{m}}\right).
	$$
\end{theorem}

\noindent{\em Remarks:} (1) The framework proves the generalization not of $f$ but of its compression $g_A$. (An exception is if the two are  shown to have similar loss at every point in the domain, not just the training set. This is the case in Theorem~\ref{thm:oldresults}.) \\
(2) The previous item highlights how our framework steps away away from uniform convergence framework, e.g., covering number arguments~\cite{dudley2010universal,anthony2009neural}. There,  one needs to fix a hypothesis class {\em independent} of the training set. By contrast we have no hypothesis class, only a {\em single} neural net that has some specific properties (described in Section~\ref{sec:stability}) on a {\em single} finite training set. But if we can compress this specific neural net to a simpler neural nets with fewer parameters then we can use covering number argument on this simpler class to get the generalization of the compressed net. \\
(3) Issue (1) exists also in standard PAC-Bayes framework for deep nets (see tongue-in-cheek title of~~\citep{langford2001not}). They yield generalization bounds not for $f$ but for a noised version of $f$ (i.e., net given by $W +\eta$, where $W$ is parameter vector of $f$ and $\eta$ is a  noise vector).  
For us issue (1) could be fixed by showing that if $f$ satisfies the properties of Section~\ref{sec:stability} on training data then it satisfies them on the entire domain. This is left for future work.

\subsection{Example 1: Linear classifiers with margin}\label{sec:linear}
 
 To illustrate the above compression method and its connection to noise stability,  we use linear classifiers with high margins. Let $c \in \R^h (\|c\| = 1)$ be a classifier for binary classification whose output on input $x$ is $\mbox{sgn}(c\cdot x)$.  Let $\mathcal{D}$ be a distribution on inputs $(x,y)$ where $\|x\|=1$ and $y\in \{\pm 1\}$. Say $c$ has margin $\gamma$ if for all $(x,y)$ in the training set we have $y(c^\top x) \ge \gamma$. 
 
 If we add Gaussian noise vector $\eta$ with coordinate-wise variance $\sigma^2$ to $c$, then $\E[x\cdot(c +\eta)]$ is $c\cdot x$ and the variance is $\sigma^2$. (A similar analysis applies to noising of $x$ instead of $c$.) Thus the margin is large if and only if the classifier's output is somewhat noise-stable. 

A classifier with margin $\gamma$ can be compressed  to one that has only $O(1/\gamma^2)$ non-zero entries. For each coordinate $i$, toss a coin with $\Pr[heads]= 8c_i^2/\gamma^2$ and if it comes up heads set the coordinate to equal to $\gamma^2/8c_i$ (see Algorithm~\ref{alg:vec-compress} in supplementary material). 
This yields a vector $\hat{c}$ with only $O(1/\gamma^2)$ nonzero entries such that for any vector $u$, with reasonable probability 
$|\hat{c}^\top u - c^\top u| \le \gamma$, so $\hat{c}$ and $c$ will make the same prediction. We can then apply Theorem~\ref{thm:compress_generalize} on a discretized version of $\hat{c}$ to show that the sparsified classifier has good generalization with $O(\log d/\gamma^2)$ samples. 

This compressed classifier works correctly for a fixed input $x$ with good probability but not high probability. To fix this, one can recourse to the 
\textquotedblleft compression with fixed string\textquotedblright\ model. The fixed string is a random linear transformation. When applied to unit vector $x$, it tends to equalize all coordinates and the guarantee $|\hat{c}^\top u - c^\top u| \le \gamma$ can hold with high probability. This random linear transformation can be fixed before seeing the training data. Similar approach was discussed by \citet{blum2006random} for linear classifiers. See Section~\ref{sec:veccompress} in supplementary material for more details. 

\subsection{Example 2: Existing generalization bounds} 
\label{subsec:existing}
Our compression framework gives  easy and short proof of the generalization bounds of a recent paper;
see appendix for slightly stronger result of~\cite{bartlett2017spectrally}.

\begin{theorem}(\cite{neyshabur2017pac}) \label{thm:oldresults} For a deep net with layers $A^1, A^2, \ldots A^d$ 
and output margin $\gamma$ on a training set $S$, the generalization error can be bounded by
$$\tilde{O}\left(\sqrt{\frac{hd^2\max_{x\in S}\|x\| \prod_{i=1}^{d} \|A^i\|_2^2 \sum_{i=1}^d \frac{\|A^i\|_F^2}{\|A^i\|_2^2}}{\gamma^2m}}\right).$$

\end{theorem}
The second part of this expression ($\sum_{i=1}^d \frac{\|A^i\|_F^2}{\|A^i\|_2^2}$) is  sum of stable ranks of the layers, a natural measure of their true parameter count. The first part ($\prod_{i=1}^{d} \|A^i\|_2^2$) is related to the Lipschitz constant of the network, namely, the maximum norm of the vector it can produce if the input is a unit vector. The Lipschitz constant of
a matrix operator $B$ is just its spectral norm $\|B\|_2$. 
Since the network applies a sequence of matrix operations interspersed with ReLU, and 
ReLU is $1$-Lipschitz 
we conclude that the Lipschitz constant of the full network is at most
$\prod_{i=1}^d \|A^i\|_2.$ 

To prove Theorem~\ref{thm:oldresults} we use the following lemma to compress the matrix at each layer to a matrix of smaller rank. Since a matrix of rank $r$ can be expressed as the product of two matrices of inner dimension $r$, it has $2hr$ parameters (instead of the trivial $h^2$). (Furthermore, the parameters can be discretized via trivial rounding to get a compression with discrete parameters as needed by Definition~\ref{def:compressible}.) 

\begin{lemma} \label{lem:detcompress} For any matrix $A\in \R^{m\times n}$, let $\hat{A}$ be the truncated version of $A$ where singular values that are smaller than $\delta\|A\|_2$ are removed. 
Then $\|\hat{A}-A\|_2\leq \delta\|A\|_2$ and $\hat{A}$ has rank at most $\|A\|_F^2/(\delta^2\|A\|_2^2)$.
\end{lemma}
\begin{proof}
Let $r$ be the rank of $\hat{A}$. By construction, the maximum singular value of $\hat{A}-A$ is at most $\delta\|A\|_2$.
 Since the remaining singular values are at least $\delta\|A\|_2$, we have $\|A\|_F \ge \|\hat{A}\|_F \ge \sqrt{r} \delta\|A\|_2$.
\end{proof}
For each $i$ replace layer $i$ by its compression using the above lemma, with 
$\delta = \gamma (3\|x\|d\prod_{i=1}^d \|A^i\|_2)^{-1}$. 
How much error does this introduce at each layer and how much does it affect the output after 
passing through the  intermediate layers (and getting magnified by their Lipschitz constants)? 
Since $A -\hat{A^i}$ has spectral norm (i.e., Lipschitz constant) at most $\delta$,
the error at the output due to changing layer $i$ in isolation is at most $\delta\|x^i\|\prod_{j=1}^d \|A^j\|_2 \le \gamma/3d$.

 A simple induction (see  \cite{neyshabur2017pac} if needed) can now  show the total error incurred in all layers is bounded by $\gamma$. The generalization bound follows immediately from Theorem~\ref{thm:compress_generalize}. 

%% file: margin.tex
\section{Noise stability properties of deep nets}
\label{sec:stability}
This section introduces noise stability properties of deep nets that imply better compression (and hence generalization). They help overcome the pessimistic error analysis of our proof of Theorem~\ref{thm:oldresults}: when a layer was compressed, the resulting error was assumed to blow up in a worst-case manner according to the Lipschitz constant (namely, product of spectral norms of layers). This  hurt the amount of compression achievable. The new noise stability properties roughly amount to saying that noise injected at a layer has very little effect on the higher layers. Our formalization starts with noise sensitivity, which captures how an operator transmits noise vs signal.
\begin{definition}\label{def:noise} If $M$ is a mapping from real-valued vectors to real-valued vectors, and ${\mathcal N}$ is some noise distribution  then {\em noise sensitivity of $M$ at $x$ with respect to ${\mathcal N}$}, is 
	$$ \psi_{{\mathcal N}}(M, x) = \E_{\eta\in \mathcal{N}}\left[\frac{\|M(x+\eta\|x\|) - M(x)\|^2}{\|M(x)\|^2}\right],$$
	The {\em noise sensitivity of $M$ with respect to ${\mathcal N}$ on a set of inputs $S$}, denoted $\psi_{{\mathcal N}, S}(M)$, is the maximum of $\psi_{{\mathcal N}}(M, x)$ over all inputs $x$ in $S$.
\end{definition}

 To illustrate, we examine noise sensitivity of a matrix (i.e., linear mapping) with respect to Gaussian distribution. Low sensitivity turns out to imply that the matrix has some large singular values (i.e., low stable rank), which give directions that can preferentially carry the \textquotedblleft signal\textquotedblright $x$ whereas noise $\eta$ attenuates because it distributes uniformly across directions. 

\begin{proposition} \label{prop:noisesensitiverank}
The noise sensitivity of a matrix $M$ at any vector $x\ne 0$  with respect to Gaussian distribution ${\mathcal N}(0, I)$  is exactly $\|M\|_F^2\|x\|^2/\|Mx\|^2$, and at least its stable rank.
\end{proposition}
\begin{proof}
Using $\E[\eta\eta^\top] = I$, we bound the numerator by
\begin{small}
\begin{align*}
&\E_{\eta}[\|M(x +\eta\|x\|) - Mx\|^2] = \E_{\eta}[\|x\|^2\|M\eta\|^2]\\
&= \E_{\eta}[\|x\|^2\mbox{tr}(M\eta\eta^\top M^\top)] = \|x\|^2 \mbox{tr}(MM^\top) = \|M\|_F^2\|x\|^2.
\end{align*}
\end{small}
Thus noise sensitivity $\psi$ at $x$ is $\|M\|_F^2\|x\|^2/\|Mx\|^2$, which is at least the stable rank $\|M\|_F^2/\|M\|_2^2$ since $\|Mx\|\le \|M\|_2\|x\|$. 
\end{proof}
The above proposition suggests that if a vector $x$ is aligned to a matrix $M$ (i.e. correlated with high singular directions of $M$), then matrix $M$ becomes less sensitive to noise at $x$. This intuition will be helpful in understanding the properties we define later to formalize noise stability.

The above discussion motivates the following approach.  We compress each layer $i$ by an appropriate randomized compression algorithm, such that the noise/error in its output is \textquotedblleft Gaussian-like\textquotedblright.   If layers $i+1$ and higher have low sensitivity to this new noise, then the compression can be more extreme producing much higher noise. 
 We formalize this idea using Jacobian $J^{i,j}$, which describes instantaneous change of $M^{i,j}(x)$ under infinitesimal perturbation of $x$.  
 
 

\subsection{Formalizing Error-resilience }\label{sec:properties}

Now we formalize the error-resilience properties. Section~\ref{sec:experiments} reports empirical findings about these properties. 
The first is {\em cushion}, to be thought of roughly as reciprocal of noise sensitivity. 
 We first formalize it for single layer.
 \begin{definition}[layer cushion] \label{def:layercushion}
 The {\em layer cushion} of layer $i$ is similarly defined to be the largest number $\mu_i$ such that for any $x\in S$, $\mu_i \|A^i\|_F\|\phi(x^{i-1})\| \leq \|A^i \phi(x^{i-1})\| $. 
 \end{definition}
Intuitively, cushion considers how much smaller the output $A^i \phi(x^{i-1})$ is compared to the upper bound $\|A^i\|_F\|\phi(x^{i-1})\|$. Using argument similar to Proposition~\ref{prop:noisesensitiverank}, we can see that $1/\mu_i^2$ is equal to the noise sensitivity of matrix $A^i$ at input $\phi(x^{i-1})$ with respect to Gaussian noise $\eta \sim {\mathcal N}(0, I)$.

  Of course, for nonlinear operators the definition of error resilience is less clean.  
 Let's denote by $M^{i,j} \colon \R^{h^i} \rightarrow \R^{h^j}$ the operator corresponding to the portion of the deep net from layer $i$ to layer $j$, and by $J^{i,j}$ its Jacobian. If infinitesimal noise is injected before level $i$ then $M^{i,j}$ passes it like $J^{i,j}$, a linear operator. When the noise is small but not infinitesimal then one hopes that $M^{i,j}$ still behaves roughly  linearly (recall that ReLU nets are piecewise linear). 
 To formalize this, we define {\em Interlayer Cushion} (Definition~\ref{def:interlayercushion}) that captures the local linear approximation of the operator $M$. 

\begin{definition}[Interlayer Cushion]\label{def:interlayercushion}
	For any two layers $i\leq j$, we define the {\em interlayer cushion} $\mu_{i,j}$ as the largest number such that for any $x\in S$:
	$$
	\mu_{i,j}\|J^{i,j}_{x^i}\|_F\|x^i\| \leq  \|J^{i,j}_{x^i}x^i\|
	$$
	Furthermore, for any layer $i$ we define the {\em minimal interlayer cushion} as $\icu = \min_{i\leq j\leq d} \mu_{i,j} = \min\{1/\sqrt{h^i},\min_{i< j\leq d} \mu_{i,j}\}$\footnote{Note that $J_{x^i}^{i,i} = I$ and $\mu_{i,i} = 1/\sqrt{h^i}$}.
\end{definition}

Since $J^{i,j}_x$  is a linear transformation, a calculation similar to Proposition~\ref{prop:noisesensitiverank} shows that
its noise sensitivity at $x^i$ with respect to Gaussian distribution $\N(0,I)$ is at most $\frac{1}{\mu_{ij}^2}$. 


The next property quantifies the intuitive observation on the learned networks that for any training data, almost half of the ReLU activations at each layer are active. If the input to the activations is well-distributed and the activations do not correlate with the magnitude of the input, then one would expect that on average, the effect of applying activations at any layer is to decrease the norm of the pre-activation vector by at most some small constant factor.

\begin{definition}[Activation Contraction]\label{def:activationcontraction}
	The activation contraction $c$ is defined as the smallest number such that for any layer $i$ and any $x\in S$,
	$$
	\|\phi(x^i)\| \geq \|x^i\|/c  .
	$$
\end{definition}



We discussed how the interlayer cushion captures noise-resilience of the network if it behaves linearly, namely, when the set of activated ReLU gates does not change upon injecting noise. In general the activations do change, but the deviation from linear behavior is bounded for small noise vectors, as quantified next. 

\begin{definition}[Interlayer Smoothness]\label{def:interlayersmoothness}
	Let $\eta$ be the noise generated as a result of substituting weights in some of the layers before layer $i$ using Algorithm~\ref{alg:matrix-proj}. We define interlayer smoothness $\rho_{\delta}$ to be the smallest number such that with probability $1-\delta$ over noise $\eta$ for any two layers $i<j$ any $x\in S$:
	$$
	\|M^{i,j}(x^{i}+\eta)-J_{x^i}^{i,j}(x^{i}+\eta)\| \leq \frac{\|\eta\|\|x^{j}\|}{\rho_{\delta} \|x^{i}\|}.
	$$
\end{definition}
In order to understand the above condition, we can look at a single layer case where $j=i+1$:
\begin{align*}
	&\|M^{i,i+1}(x^{i}+\eta)-J_{x^i}^{i,i+1}(x^{i}+\eta)\|=\|A^{i+1}\phi(x^{i}+\eta)-A^{i+1} (\phi'(x^i) \odot (x^{i}+\eta))\|\\
	&=\|A^{i+1} \nu\| \leq \frac{\|\eta\|\|A^{i+1} \phi(x^i)\|}{\rho_{\delta} \|x^{i}\|}
\end{align*}
where $\odot$ is the entry-wise product operator and $\nu = (\phi'(x^{i}+\eta) - \phi'(x^{i})) \odot (x^{i}+\eta)$. Since the activation function is ReLU, $\phi'(x^{i}+\eta)$ and $\phi'(x^{i}))$ disagree whenever the perturbation has the opposite sign and higher absolute value compared to the input and hence $\|\nu\|\leq \|\eta\|$.

Let us first see what happens if the perturbation $\nu$ is adversarially aligned to the weights:
\begin{align*}
&\|M^{i,i+1}(x^{i}+\eta)-J_{x^i}^{i,i+1}(x^{i}+\eta)\|\\ &=\|A^{i+1} \nu\| \leq \|A^{i+1}\|\|\eta\|= \frac{\|\eta\|\|A^{i+1} \phi(x^i)\|}{\|x^{i}\|} \cdot\frac{\|A^{i+1}\|\|x^{i}\|}{\|A^{i+1} \phi(x^i)\|}\\
&\leq \frac{\|\eta\|\|A^{i+1} \phi(x^i)\|}{\|x^{i}\|} \cdot\frac{\|A^{i+1}\|\|x^{i}\|}{\mu_{i+1}\|A^{i+1}\|_F\|\phi(x^i)\|}\leq \frac{\|\eta\|\|A^{i+1} \phi(x^i)\|}{\|x^{i}\|} \cdot\frac{c\|A^{i+1}\|}{\mu_{i+1}\|A^{i+1}\|_F}\\
&= \frac{\|\eta\|\|A^{i+1} \phi(x^i)\|}{\|x^{i}\|} \cdot\frac{c}{\mu_{i+1} r_{i+1}}\\
\end{align*}
where $r_{i+1}$ is the stable rank of layer $i+1$. Therefore the interlayer smoothness from layer $i$ to layer $i+1$ is at least $\rho_\delta = \mu_{i+1} r_{i+1}/c$. However, the noise generated from Algorithm~\ref{alg:matrix-proj} has similar properties to Gaussian noise (see Lemma~\ref{lem:perturblayer:highprob}). If $\nu$ behaves similar to Gaussian noise, then $\|A^{i+1}\nu\| \approx \|A^{i+1}\|_F\|\nu\|/\sqrt{h^{i}}$ and therefore $\rho_\delta$ is as high as $\sqrt{h^{i}\mu_{i+1}/c}$. Since the layer cushion of networks trained on real data is much more than that of networks with random weights, $\rho_\delta$ is greater than one in this case. Another observation is that in practice, the noise is well-distributed and only a small portion of activations change from active to inactive and vice versa. Therefore, we can expect $\|\nu\|$ to be smaller than $\|\eta\|$ which further improves the interlayer smoothness. This appeared in \cite{neyshabur2017exploring} that showed for one layer we can even use $\frac{\|\eta\|^{1.5}\|x^{j}\|}{\rho_{\delta} \|x^{i}\|}$ as the RHS of interlayer smoothness. Our current proof requires $1/\rho_\delta$ to be of order $1/d$, this requirement can be removed (with $\rho_\delta$ appear in sample complexity) if we make the stronger assumption that the RHS is a lower order term in $\|\eta\|$.

In general, for a single layer, $\rho_\delta$ captures the ratio of input/weight alignment to noise/weight alignment. Since the noise behaves similar to Gaussian, one expects this number to be greater than one for a single layer. When $j>i+1$, the weights and activations create more dependencies. However, since these dependences are applied on both noise and input, we again expect that if the input is more aligned to the weights than noise, this should not change in higher layers. In Section~\ref{sec:experiments}, we show that the interlayer smoothness is indeed good: $1/\rho_\delta$ is a small constant.

%% file: newmatrix.tex
\section{Fully Connected Networks}
\label{sec:newmatrix}

We prove generalization bounds using for fully connected multilayer nets. Details appear in  Appendix Section~\ref{appx:newmatrix}.

\begin{theorem}\label{thm:fullyconnected_generalization}
	For any fully connected network $f_A$ with $\rho_\delta\geq 3d$, any probability $0<\delta\leq 1$ and any margin $\gamma$, Algorithm~\ref{alg:matrix-proj} generates weights $\tilde{A}$ for the network $f_{\tilde{A}}$ such that with probability $1-\delta$ over the training set and $f_{\tilde{A}}$, the expected error $L_0(f_{\tilde{A}})$ is bounded by
	\begin{equation*}
	 \hat{L}_\gamma(f_A) + \tilde{O}\left(\sqrt{\frac{c^2d^2\max_{x\in S}\|f_A(x)\|_2^2 \sum_{i=1}^d \frac{1}{\mu_i^2\icu^2}}{\gamma^2 m}}\right)
	\end{equation*}
	where $\mu_i$, $\icu$, $c$ and $\rho_{\delta}$ are layer cushion, interlayer cushion, activation contraction and interlayer smoothness defined in Definitions~\ref{def:layercushion},\ref{def:interlayercushion},\ref{def:activationcontraction} and \ref{def:interlayersmoothness} respectively.
\end{theorem}



 To prove this we describe a compression of the net with respect to a fixed (random) string. In contrast to the deterministic compression of Lemma~\ref{lem:detcompress}, this  randomized compression ensures that the resulting error in the output behaves like a Gaussian. The proofs are similar to standard JL dimension reduction. 

\begin{algorithm}
	\caption{Matrix-Project ($A$, $\eps$, $\eta$)}
	\begin{algorithmic}
		\REQUIRE Layer matrix $A\in \R^{h_1\times h_2}$, error parameter $\eps$, $\eta$.
		\ENSURE Returns $\hat{A}$ s.t. $\forall$ fixed vectors $u,v$, $$\Pr[|u^\top \hat{A} v - u^\top Av\| \ge \eps\|A\|_F\|u\|\|v\|] \le \eta.$$
		\STATE Sample $k = \log(1/\eta)/\eps^2$ random matrices $M_1,\dots,M_k$ with entries i.i.d. $\pm 1$ (\textquotedblleft helper string\textquotedblright)
		\FOR{$k'=1$ to $k$}
		\STATE Let $Z_{k'} = \inner{A,M_{k'}} M_{k'}$.
		\ENDFOR
		\STATE Let $\hat{A} =\frac{1}{k}\sum_{k'=1}^k Z_{k'}$
	\end{algorithmic}
	\label{alg:matrix-proj}
\end{algorithm}
Note that the helper string of random matrices $M_i$'s were chosen and fixed before training set $S$ was picked. 
Each weight matrix is thus represented as only $k$ real numbers $\inner{A,M_i}$ for $i = 1,2,...,k$. 

\begin{lemma}\label{lem:perturblayer:highprob}
For any $0<\delta,\eps\leq 1$, et $G=\{(U^i,x^i)\}_{i=1}^m$ be a set of matrix/vector pairs of size $m$ where $U\in\R^{n\times h_1}$ and $x\in\R^{h_2}$, let $\hat{A}\in \R^{h_1\times h_2}$ be the output of Algorithm~\ref{alg:matrix-proj} with $\eta = \delta/mn$. With probability at least $1-\delta$ we have for any $(U,x)\in G$, $\|U (\hat{A} - A) x\| \leq \eps\|A\|_F\|U\|_F\|x\|$.
\end{lemma}

\begin{remark}
	Lemma~\ref{lem:perturblayer:highprob} can be used to upper bound the change in the network output after compressing a single layer if the activation patterns remain the same. For any layer, in the lemma statement take $x$ to be the input to the layer, $A$ to be the layer weight matrix, and $U$ to be the Jacobian of the network output with respect to the layer output.  Network output before and after compression can then be calculated by the matrix products $UAx$ and $U\hat{A}x$ respectively. Hence, the lemma bounds the distance between network output before and after compression.
\end{remark}
Next Lemma bounds the number of parameters of the compressed network resulting from applying Algorithm ~\ref{alg:matrix-proj} to all the layer matrices of the net.  The proof does induction on the layers and bounds the effect of the error on the output of the network using properties defined in Section~\ref{sec:properties}. 

\begin{lemma}\label{lem:rotate-compress}
	For any fully connected network $f_A$ with $\rho_\delta\geq 3d$, any probability $0<\delta\leq 1$ and any error $0<\eps\leq 1$, Algorithm~\ref{alg:matrix-proj} generates weights $\tilde{A}$ for a network with $\frac{72c^2d^2\log (mdh/\delta)}{\eps^2}\cdot \sum_{i=1}^d \frac{1}{\mu_i^2\icu^2}$ total parameters such that with probability $1-\delta/2$ over the generated weights $\tilde{A}$, for any $x\in S$:
	$$
	\|f_A(x) - f_{\tilde{A}}(x)\| \le \eps\|f_A(x)\|.
	$$
	where $\mu_i$, $\icu$, $c$ and $\rho_{\delta}$ are layer cushion, interlayer cushion, activation contraction and interlayer smoothness defined in Definitions~\ref{def:layercushion},\ref{def:interlayercushion},\ref{def:activationcontraction} and \ref{def:interlayersmoothness} respectively.
\end{lemma}

\paragraph{Some obvious improvements:} 
 (i) Empirically it has been observed that deep net training introduces fairly small changes to parameters as compared to the (random) initial weights~\cite{dziugaite2017computing}. We can exploit this by incorporating the random initial weights into the helper string and  do the entire proof above not with the layer matrices $A^i$ but only the difference from the initial starting point.
 Experiments in Section~\ref{sec:experiments} show this improves the bounds. 
(ii) Cushions and other quantities defined earlier are  data-dependent, and required to hold for the {\em entire} training set. However, the proofs go through if we remove say $\zeta$ fraction of outliers that violate the definitions; this allows us to
use more favorable values for cushion etc. and lose an additive factor $\zeta$ in the generalization error. 

%% file: convolutionmain.tex
\section{Convolutional Neural Networks}
\label{sec:convolution}
Now we sketch how to provably compress convolutional nets. (Details appear in Section~\ref{sec:convolution:appendix} of supplementary.) Intuitively, this feels harder  because the weights are already compressed--- they're {\em shared} across patches! 

\begin{theorem}\label{thm:convolution:main}
	For any convolutional neural network $f_A$ with $\rho_\delta\geq 3d$, any probability $0<\delta\leq 1$ and any margin $\gamma$, Algorithm~\ref{alg:matrix-proj-p} generates weights $\tilde{A}$ for the network $f_{\tilde{A}}$ such that with probability $1-\delta$ over the training set and $f_{\tilde{A}}$:
\begin{align*}
&L_0(f_{\tilde{A}}) \leq \hat{L}_\gamma(f_A) 
\\&+ \tilde{O}\left(\sqrt{\frac{c^2d^2 \max_{x\in S}\|f_A(x)\|_2^2 \sum_{i=1}^d \frac{\beta^2(\lceil\kappa_i/s_i\rceil)^2}{\mu_i^2\icu^2}}{\gamma^2 m}}\right)
\end{align*}
where $\mu_i$, $\icu$, $c$, $\rho_{\delta}$ and $\beta$ are layer cushion, interlayer cushion, activation contraction, interlayer smoothness and well-distributed Jacobian defined in Definitions~\ref{def:layercushion},\ref{def:interlayercushionLconv},\ref{def:activationcontraction}, \ref{def:interlayersmoothness} and \ref{def:welldistributed:jacobian} respectively. Furthermore, $s_i$ and $\kappa_i$ are stride and filter width in layer $i$.
\end{theorem}
Let's realize that obvious extensions of earlier sections fail. Suppose layer $i$ of the neural network is an image of dimension $n^i_1\times n^i_2$ and each pixel has $h^i$ channels, the size of the filter at layer $i$ is $\kappa_i\times \kappa_i$ with stride $s_i$. The convolutional filter has dimension $h^{i-1}\times h^{i}\times \kappa_i\times \kappa_i$. Applying matrix compression (Algorithm 1) independently to {\em each copy} of a convolutional filter 
makes number of new parameters proportional to $n^i_1n^i_2$, a big blowup.

Compressing a convolutional filter once and reusing it  in all patches doesn't work because the interlayer analysis implicitly requires the noise generated by the compression to behave similar to a spherical Gaussian, but the shared filters introduce correlations. Quantitatively, using the fully connected analysis would require the error to be less than
interlayer cushion value $\mu_{i\to}$ (Definition~\ref{def:interlayercushion}) which is at most $1/\sqrt{h^in^i_1n^i_2}$, and this can never be achieved from compressing matrices that are far smaller than $n^i_1 \times n^i_2$  to begin with.

We end up with a solution  in between fully independent and fully dependent: {\em p-wise} independence.
The algorithm  generates $p$-wise independent compressed filters $\hat{A}_{(a,b)}$ for each convolution location $(a,b)\in[n^i_1]\times [n^i_2]$. It results in $p$ times more parameters than a single compression. If $p$ grows logarithmically with relevant parameters,  the filters behave like fully independent filters. Using this idea we can generalize the definition of interlayer margin to the convolution setting:


\begin{definition}[Interlayer Cushion, Convolution Setting]\label{def:interlayercushionLconv}
	For any two layers $i\leq j$, we define the {\em interlayer cushion} $\mu_{i,j}$ as the largest number such that for any $x\in S$:
	$$
	\mu_{i,j}\cdot \frac{1}{\sqrt{n^i_1n^i_2}}\|J^{i,j}_{x^i}\|_F\|x^i\| \leq  \|J^{i,j}_{x^i}x^i\|
	$$
	Furthermore, for any layer $i$ we define the {\em minimal interlayer cushion} as $\icu = \min_{i\leq j\leq d} \mu_{i,j} = \min\{1/\sqrt{h^i},\min_{i< j\leq d} \mu_{i,j}\}$\footnote{Note that $J_{x^i}^{i,i} = I$ and $\mu_{i,i} = 1/\sqrt{h^i}$}.
\end{definition}


Recall that interlayer cushion is related to the noise sensitivity of $J^{i,j}_{x^i}$ at $x^i$ with respect to Gaussian distribution $\N(0,I)$. When we consider $J^{i,j}_{x^i}$ applied to a noise $\eta$, if different pixels in $\eta$ are independent Gaussians, then we can indeed expect $\|J^{i,j}_{x^i} \eta\| \approx \frac{1}{\sqrt{h^in^i_1n^i_2}}\|J^{i,j}_{x^i}\|\|\eta\|$, which explains the extra $\frac{1}{\sqrt{n^i_1n^i_2}}$ factor in Definition~\ref{def:interlayercushionLconv} compared to Definition~\ref{def:interlayercushion}.
The proof also needs to assume ---in line with intuition behind convolution architecture--- that information from the entire image field is incorporated somewhat uniformly across pixels. It is formalized using the Jacobian which gives the partial derivative of the output with respect to pixels at previous layer.  

\begin{definition}[Well-distributed Jacobian]\label{def:welldistributed:jacobian} Let $J^{i,j}_x$ be the Jacobian of $M^{i,j}$ at $x$, we know $J^{i,j}_x \in \R^{h^i\times n^i_1\times n^i_2\times h^j\times n^j_1\times n^j_2}$. We say the Jacobian is $\beta$ well-distributed if for any $x\in S$, any $i,j$, any $(a,b)\in[n^i_1\times n^i_2]$,
$$
\|[J^{i,j}_x]_{:,a,b,:,:,:}\|_F \le \frac{\beta}{\sqrt{n^i_1n^i_2}}\|J^{i,j}_x\|_F
$$
\end{definition}

%% file: experiments.tex
\section{Empirical Evaluation}
\label{sec:experiments}
We study noise stability properties (defined in Section~\ref{sec:stability}) of an actual trained deep net, and compute a generalization bound from Theorem~\ref{thm:convolution:main}. Experiments were performed by training a VGG-19 architecture \cite{simonyan2014very} and a AlexNet \cite{krizhevsky2012imagenet} for multi-class classification task on CIFAR-10 dataset. Optimization used SGD with mini-batch size 128, weight decay 5e-4,  momentum $0.9$ and initial learning rate $0.05$, but decayed by factor 2 every 30 epochs.  Drop-out was used in fully-connected layers. We trained both networks for 299 epochs and the final VGG-19 network achieved $100\%$ training and $92.45\%$ validation accuracy while the AlexNet achieved $100\%$ training and $77.22\%$ validation accuracy. To investigate the effect of corrupted label, we trained another AlexNet, with $100\%$ training and $9.86\%$ validation accuracy, on CIFAR-10 dataset with randomly shuffled labels.

Our estimate of the sample complexity bound used exact computation of norms of weight matrices (or tensors) in all bounds($||A||_{1,\infty}, ||A||_{1,2}, ||A||_{2}, ||A||_F$). Like previous bounds in generalization theory, ours also depend upon nuisance parameters like  depth $d$, logarithm of $h$, etc.  which probably are an artifact of the proof. These are ignored in the computation (also in computing earlier bounds) for simplicity. 
Even the generalization based on parameter counting arguments does have an extra dependence on depth~\cite{bartlett2017spectrally}. A recent work, \cite{golowich2017size} showed that many such depth dependencies can be improved.

\subsection{Empirical investigation of noise stability properties}\label{exp:verify_assumption}
\begin{figure}[ht]
    \centering
    \includegraphics[width=10cm]{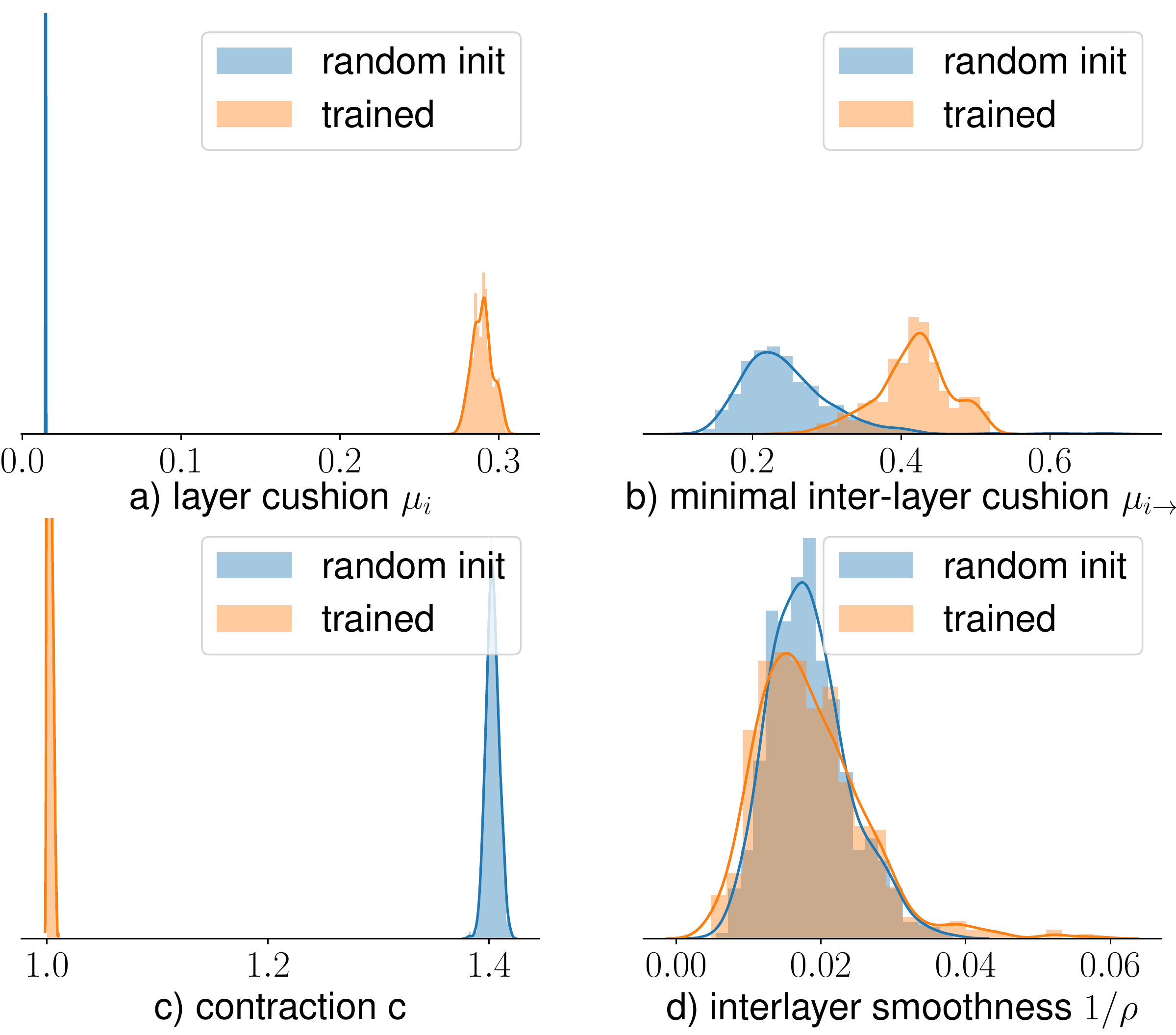}
    \caption{Distribution of a) layer cushion, b) (unclipped) minimal interlayer cushion, c) activation contraction and d) interlayer smoothness of the 13-th layer of VGG-19 nets on on training set. The distributions on a randomly-initialized and a trained net are shown in blue and orange. Note that after clipping, the minimal interlayer cushion is set to $1/\sqrt{h_i}$ for all layers except the first one, see appendix~\ref{appdix:verification}.}
    \label{fig:verify}
    \vspace{-0.5cm}
\end{figure}

Section \ref{sec:stability}  identifies four properties in the networks that contribute to noise-stability: layer cushion, interlayer cushion, contraction, interlayer smoothness. Figure~\ref{fig:verify} plots the distribution of  over different data points in the training set and compares  to a Gaussian random network and then scaled properly. The layer cushion, which quantifies its noise stability, is drastically improved during the training, especially for the higher layers ($8$ and higher) where most parameters live. Moreover, we observe that interlayer cushion, activation contraction and interlayer smoothness behave nicely even after training.  These plots suggest that the  driver of the generalization phenomenon is layer cushion.  The other properties are  being maintained in the network and prevent the network from falling prey to pessimistic assumptions that causes the other older generalization bounds to be very high.
The assumptions made in section~\ref{sec:stability} (also in~\ref{conditions}) are verified on the VGG-19 net in appendix~\ref{appdix:verification} by histogramming the distribution of layer cushion, interlayer cushion, contraction, interlayer smoothness, and well-distributedness of the Jacobians of each layer of the net on each data point in the training set. Some examples are shown in Figure~\ref{fig:verify}.

\subsection{Correlation to generalization error}
We evaluate our generalization bound during the training, see Figure~\ref{fig:comparison}, Right. 
After 120 epochs, the training error is almost zero but the test error continues to improve in later epochs. Our generalization bound continues to improve, though not to the same level. Thus our generalization bound captures part of generalization phenomenon, not all.
Still, this suggests that SGD somehow improves our generalization measure implicitly. Making this rigorous is a good topic for further research.

Furthermore, we investigate effect of training with normal data and corrupted data by training two AlexNets respectively on original and corrupted CIFAR-10 with randomly shuffled labels. We identify two key properties that differ significantly between the two networks: layer cushion and activation contraction, see~\ref{fig:comparison}. Since our bound predicts larger cushion and lower contraction indicates better generalization, our bound is consistent w with the fact that the net trained on normal data generalizes ($77.22\%$ validation accuracy).


\subsection{Comparison to other generalization bounds}\label{sec:comparison}
\begin{figure}[H]
    \centering
    \begin{subfigure}{0.45\textwidth}
        \centering
        \includegraphics[width=5cm,height=4cm]{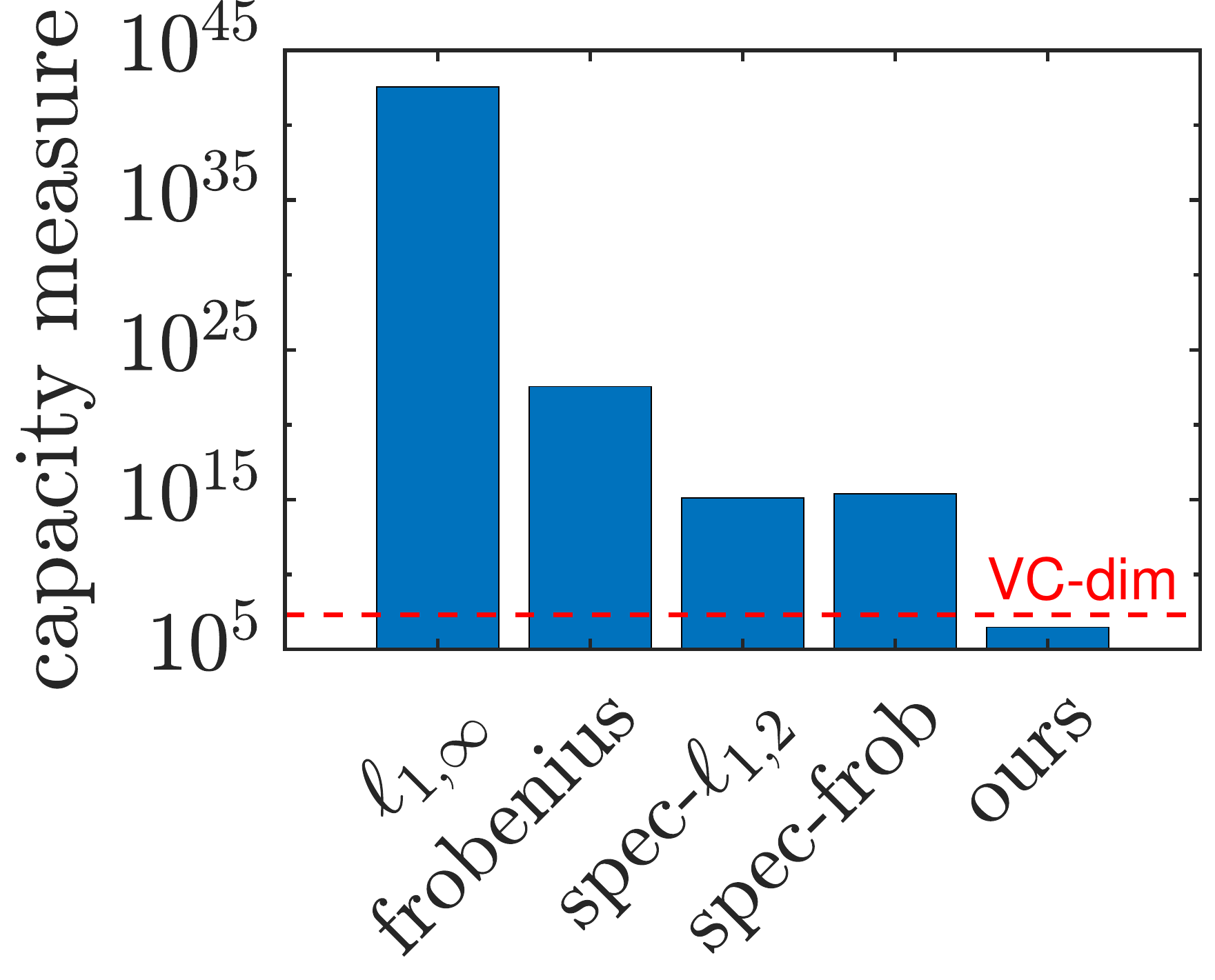}
    \end{subfigure}
    ~
    \begin{subfigure}{0.45\textwidth}
        \centering
        \vspace{-0.3cm}
        \includegraphics[width=5cm, height=3.5cm]{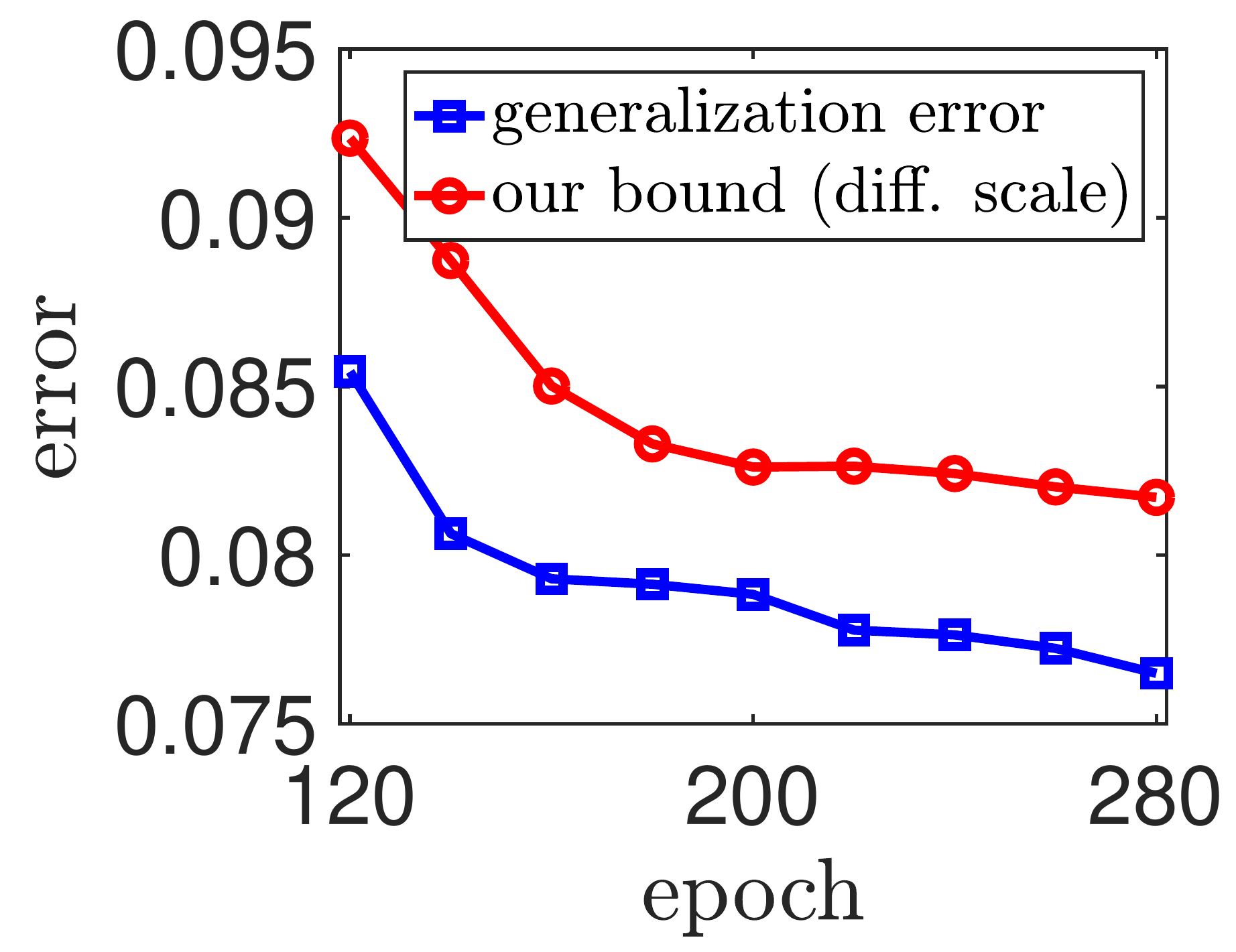}
    \end{subfigure}
    \vspace{-0.2cm}
    \caption{\textbf{Left)} Comparing neural net genrealization bounds. \\
        $\ell_{1,\infty}:$ $\frac{1}{\gamma^2}\prod_{i=1}^d||A^i||_{1,\infty}$~\cite{bartlett2002rademacher}\\
        Frobenius: $\frac{1}{\gamma^2}\prod_{i=1}^d||A^i||_F^2$~\cite{NeyTomSre15},\\
        spec $\ell_{1,2}$: $\frac{1}{\gamma^2}\prod_{i=1}^d||A_i||_2^2\sum_{i=1}^d\frac{||A^i||_{1,2}^2}{||A^i||_2^2}$ \cite{bartlett2017spectrally}\\
        spec-fro: $\frac{1}{\gamma^2}\prod_{i=1}^d||A^i||_2^2\sum_{i=1}^dh_i\frac{||A^i||_F^2}{||A^i||_2^2}$~\cite{neyshabur2017pac}\\
        ours: $\frac{1}{\gamma^2}\max_{x\in S}||f(x)||_2^2\sum_{i=1}^d\frac{\beta^2 c_i^2 \lceil\kappa/s\rceil^2 }{\mu_i^2\mu_{i\rightarrow}^2}$\\
    \textbf{Right)} Comparing our bound to empirical generalization error during training. Our bound is rescaled to be within the same range as the generalization error.}
    \label{fig:comparison}
    \vspace{-0.5cm}
\end{figure}

Figure~\ref{fig:comparison} compares our proposed bound to other neural-net generalization bounds on the VGG-19 net and compares to naive VC dimension bound (which of course is too pessimistic).  All previous generalization bounds are orders of magnitude worse than ours; the closest one is spectral norms times average $\ell_{1,2}$ of the layers \cite{bartlett2017spectrally} which is still about $10^{15}$, far greater than VC dimension. (As mentioned we're ignoring nuisance factors like depth and $\log h$ which make the comparison to VC dimension a bit unfair, but the comparison to previous bounds is fair.)
This should not be surprising as all other bounds are based on product of norms which is pessimistic (see note at the start of Section~\ref{sec:stability}) which we avoid due to the  noise stability analysis.

Table~\ref{table:compress1} shows the compressibility of various layers according to the bounds given by our theorem. Again, this is a qualitative due to ignoring nuisance factors, but it gives an idea of which layers are important in the calculation.

\begin{figure}[H]
    
    \centering

        \includegraphics[width=10cm, height=3.5cm]{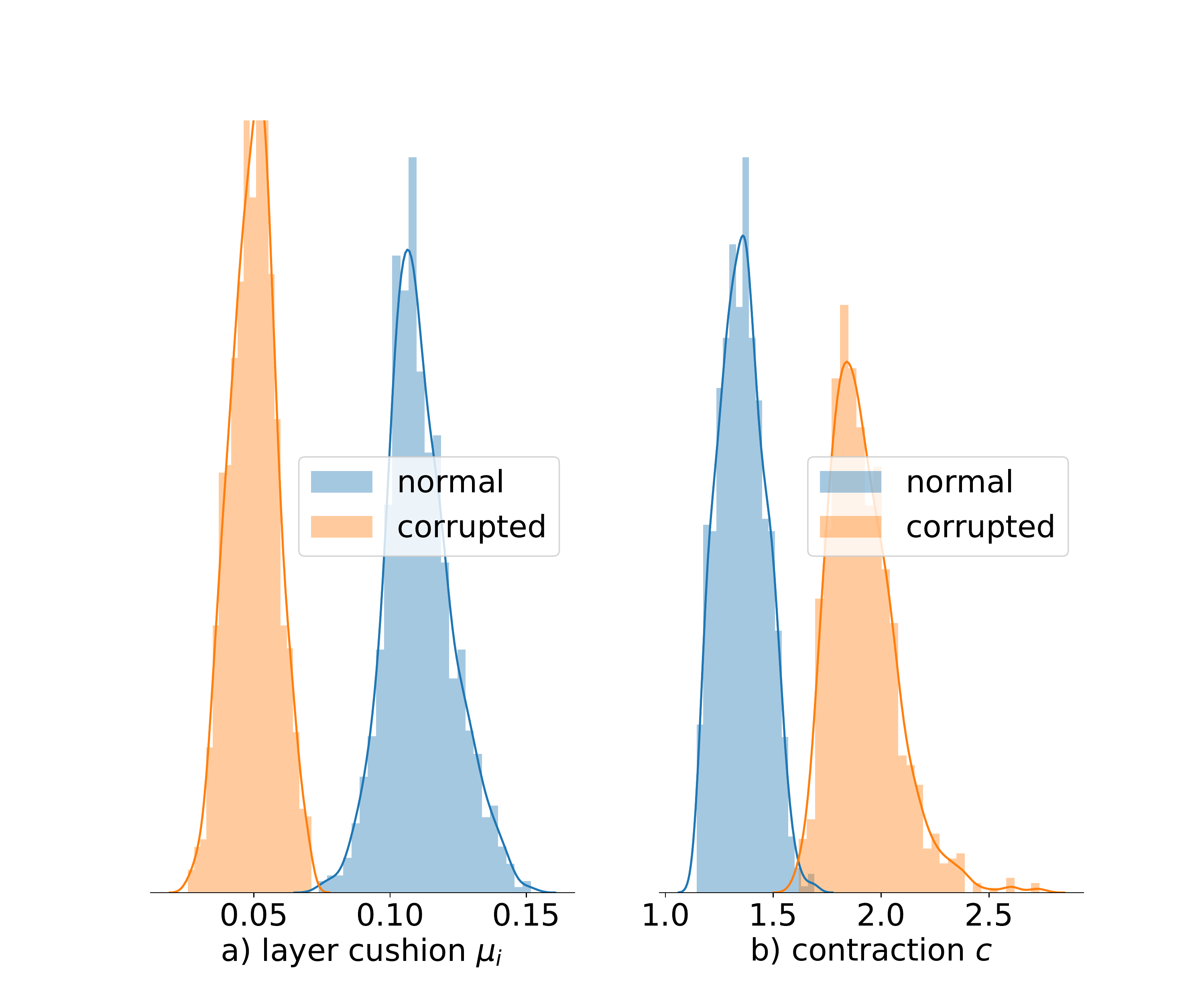}
        \caption{Distribution of a) layer cushion, b) activation contraction on training set of the 5-th layer of AlexNets nets trained on normal and corrupted dataset. The distributions of the two nets are shown in blue and orange. Note that the net trained on normal data (blue) has lager layer cushion and smaller activation contraction. For statistics of other layers, see~\ref{appendix:corrupted}}

    
\end{figure}

\begin{table}[H]
\centering
 \begin{tabular}{c | c c c} 

 layer & $  \frac{c_i^2\beta_i^2\lceil\kappa_i/s_i\rceil^2}{\mu_i^2\mu_{i\rightarrow}^2} $ &  actual \# param & compression ($\%$) \\ [0.5ex] 
 \hline
  1 & 1644.87 & 1728 & 95.18 \\ 
  \hline
 4 & 644654.14 & 147456 & 437.18\\
 \hline
  6 & 3457882.42 & 589824 & 586.25\\
  \hline
 9 & 36920.60 & 1179648 & 3.129 \\
  \hline 
 12 & 22735.09  & 2359296 & 0.963\\
 \hline
 15 & 26583.81 & 2359296 & 1.126 \\
 \hline
 18 & 5052.15 & 262144 & 1.927  \\

\end{tabular}
\caption{Effective number of parameters identified by our bound. Compression rates can be as low as $1\%$ in later layers (from 9 to 19) whereas earlier layers are not so compressible. Dependence on depth $d$, log factors, constants are ignored  as mentioned in the text.}\label{table:compress1}
\end{table}

%% file: conclusions.tex
\section{Conclusions}
With a new compression-based approach, the paper has made progress on several open issues regarding generalization properties of 
deep nets. The approach also adapts specially to convolutional nets. 
The empirical verification of the theory in Section~\ref{sec:experiments} shows a rich set of new properties satisfied by deep nets trained on realistic data, which we hope will fuel further theory work on deep learning, including how these properties play into optimization and expressivity. Another possibility is a more rigorous understanding of deep net compression, which sees copious empirical work motivated by low-power applications.  Perhaps our p-wise independence idea used for compressing convnets (Section~\ref{sec:convolution}) has practical implications.

%% file: appendix.tex
\appendix 
\section{Complete proofs for Section~\ref{sec:compress}}

In this section we give proofs of the various statements.

\subsection{Generalization Bounds from Compression}

We will first prove Theorem~\ref{thm:compress_generalize}, which gives generalization guarantees for the compressed function.

\begin{proof}(Theorem~\ref{thm:compress_generalize})
The proof is a basic application of concentration bounds and union bound and appears in the appendix. 

For each $A\in \mathcal{A}$, the training loss $\hat{L}_0(g_A)$ is just the average of $n$ i.i.d. Bernoulli random variables with expectation equal to $L_0(g_A)$. Therefore by Chernoff bound we have
$$
\Pr[L_0(g_A) - \hat{L}_0(g_A) \ge \tau] \le \exp(-2\tau^2 n).
$$

Therefore, suppose we choose $\tau = \left(\sqrt{\frac{m\log q}{n}}\right)$, with probability at least $1-\exp(-2m\log q)$ we have $L_0(g_A) \le \hat{L}_0(g_A) + \tau$. There are only $q^m$ different $A \in \mathcal{A}$, hence by union bound, with probability at least $1-\exp(-m\log q)$, for all $A\in \mathcal{A}$ we have
$$L_0(g_A) \le \hat{L}_0(g_A) + \left(\sqrt{\frac{m\log q}{n}}\right).$$

Next, since $f$ is $(\gamma,S)$-compressible with respect to $g$, there exists $A\in \mathcal{A}$ such that for $x\in S$ and any $y$ we have
$$
|f(x)[y] - g_A(x)[y]| \le \gamma.
$$
For these training examples, as long as the original function $f$ has margin at least $\gamma$, the new function $g_A$ classifies the example correctly. Therefore
$$\hat{L}_0(g_A) \le \hat{L}_\gamma(f).$$
Combining these two steps, we immediately get the result.
\end{proof}

%

Using the same approach, we can also prove the following Corollaries that allow the compression to fail with some probability

\begin{corollary}\label{cor:compression}
In the setting of Theorem~\ref{thm:compress_generalize}, if the compression works for $1-\zeta$ fraction of the training sample, then with high probability
$$
	L_0(g_A) \le \hat{L}_\gamma(f)+ \zeta+O\left(\sqrt{\frac{q\log r}{m}}\right).
	$$
%
	\end{corollary}

\begin{proof}
The proof is using the same approach, except in this case  we have
$$
\hat{L}_0(g_A) \le \hat{L}_\gamma(f) + \zeta.
$$
\end{proof}

\subsection{Example 1: Compress a Vector}
\label{sec:veccompress}

This section gives detailed calculations supporting the first example in Section~\ref{sec:compress}.

\begin{algorithm}
\caption{Vector-Compress($\gamma$, $c$)}
\begin{algorithmic}
\REQUIRE vector $c$ with $\|c\| \le 1$, $\eta$.
\ENSURE vector $\hat{c}$ s.t. for any fixed vector $\|u\| \le 1$, with probability at least $1-\eta$, $|c^\top u - \hat{c}^\top u| \le \gamma$. Vector $\hat{c}$ has $O((\log h)/\eta\gamma^2)$ nonzero entries. 
\FOR{$i = 1$ to $d$}
\STATE Let $z_i = 1$ with probability $p_i = \frac{2c_i^2}{\eta\gamma^2}$ (and $0$ otherwise)
\STATE Let $\hat{c}_i = z_i c_i/p_i$.
\ENDFOR
\STATE Return $\hat{c}$
\end{algorithmic}
\label{alg:vec-compress}
\end{algorithm}

\begin{lemma}
\label{lem:vec-compress}
Algorithm~\ref{alg:vec-compress} Vector-Compress($\gamma$, $c$) returns a vector $\hat{c}$ such that for any fixed $u$ (independent of choice of $\hat{c}$), with probability at least $1-\eta$, $|\hat{c}^\top u - c^\top u| \le \gamma$. The vector $\hat{c}$ has at most $O((\log h)/\eta\gamma^2)$ non-zero entries with high probability.
\end{lemma}

\begin{proof}
By the construction in Algorithm~\ref{alg:vec-compress}, it is easy to check that for all $i$, $\E[\hat{c}_i] = c_i$. Also, $\Var[\hat{c}] = 2p_i(1-p_i) \frac{c_i^2}{p_i^2} \le \frac{2c_i^2}{p_i} \le \eta\gamma^2$.

Therefore, for any vector $u$ that is independent with the choice of $\hat{c}$, we have $\E[\hat{c}^\top u] = c^\top u$ and $\Var[\hat{c}^\top u] \le \|u\|^2/4 \le \eta\gamma^2$. Therefore by Chebyshev's inequality we know $\Pr[|\hat{c}^\top u - c^\top u| \ge \gamma] \le \eta$.

On the other hand, the expected number of non-zero entries in $\hat{c}$ is $\sum_{i=1}^d p_i = 2/\eta\gamma^2$. By Chernoff bound we know with high probability the number of non-zero entries is at most $O((\log h)/\eta\gamma^2)$.
\end{proof}

Next we handle the discretization:

\begin{lemma}\label{lem:discretizevec}
Let $\tilde{c} = \mbox{Vector-Compress}(\gamma/2, c)$. For each coordinate $i$, let $\hat{c}_i = 0$ if $|\tilde{c}_i| \ge 2\eta\gamma\sqrt{h}$, otherwise let $\hat{c}_i$ be the rounding of $\tilde{c}_i$ to the nearest multiple of $\gamma/2\sqrt{h}$. For any fixed $u$  with probability at least $1-\eta$, $|\hat{c}^\top u - c^\top u| \le \gamma$.
\end{lemma}

\begin{proof}
Let $c'$ be a truncated version of $c$: $c'_i = c_i$ if $|c_i| \ge \gamma/4\sqrt{h}$, and $c'_i = 0$ otherwise. It is easy to check that $\|c'-c\| \le \gamma/4$. By Algorithm~\ref{alg:vec-compress}, we observe that $\tilde{c} = \mbox{Vector-Compress}(\gamma/2, c')$ ($|\tilde{c}_i| \ge 2\eta\gamma\sqrt{h}$ if and only if $|c_i| \le \gamma/4\sqrt{h}$). Finally, by the rounding we know $\|\hat{c}-\tilde{c}\|\le \gamma/4$. Combining these three terms, we know with probability at least $1-\eta$,
\begin{align*}
|\hat{c}^\top u - c^\top u| &\le  |\hat{c}^\top u - \tilde{c}^\top u|+|\tilde{c}^\top u - (c')^\top u|+|(c')^\top u - c^\top u| \\
& \le \gamma/4+\gamma/2+\gamma/4 = \gamma.
\end{align*}
\end{proof}

Combining the above two lemmas, we know there is a compression algorithm with $O((\log h)/\eta\gamma^2)$ discrete parameters that works with probability at least $1-\eta$. Applying Corollary~\ref{cor:compression} we get

\begin{lemma}
For any number of sample $m$, there is an efficient algorithm to generate a compressed vector $\hat{c}$, such that 
$$
L(\hat{c}) \le \tilde{O}((1/\gamma^2 m)^{1/3}).
$$
\end{lemma}

\begin{proof}
We will choose $\eta = (1/\gamma^2 m)^{1/3}$. By Lemma~\ref{lem:vec-compress} and Lemma~\ref{lem:discretizevec}, we know there is a compression algorithm that works with probability $1-\eta$, and has at most $\tilde{O}((\log h)/\eta\gamma^2)$ parameters. By Corollary~\ref{cor:compression}, we know
$$
L(\hat{c}) \le \tilde{O}(\eta + \sqrt{1/\eta\gamma^2m}) \le \tilde{O}((1/\gamma^2 m)^{1/3}).
$$
\end{proof}

Note that the rate we have here is not optimal as it depends on $m^{1/3}$ instead of $\sqrt{m}$. This is mostly due to Lemma~\ref{lem:vec-compress} cannot give a high probability bound (indeed if we consider all the basis vectors as the test vectors $u$, Vector-Compress is always going to fail on some of them). 

\paragraph{Compression with helper string} To fix this problem we use a different algorithm that uses a helper string, see Algorithm~\ref{alg:vec-proj}

\begin{algorithm}
\caption{Vector-Project($\gamma$, $c$)}
\begin{algorithmic}
\REQUIRE vector $c$ with $\|c\| \le 1$, $\eta$.
\ENSURE vector $\hat{c}$ s.t. for any fixed vector $\|u\| \le 1$, with probability at least $1-\eta$, $|c^\top u - \hat{c}^\top u| \le \gamma$. 
\STATE Let $k = 16\log(1/\eta)/\gamma^2$
\STATE Sample $k$ random Gaussian vectors $v_1,...,v_k\sim \N(0,I)$.
\STATE Compute $z_i = \inner{v_i,c}$
\STATE (Optional): Round $z_i$ to the closes multiple of $\gamma/2\sqrt{hk}$.
\STATE Return $\hat{c} = \frac{1}{k}\sum_{i=1}^k z_i v_i$
\end{algorithmic}
\label{alg:vec-proj}
\end{algorithm}

Note that in Algorithm~\ref{alg:vec-proj}, the parameters for the output are the $z_i$'s. The vectors $v_i$'s are sampled independently, and hence can be considered to be in the helper string.

\begin{lemma}\label{lem:vec-proj}
For any fixed vector $u$, Algorithm~\ref{alg:vec-proj} $\mbox{Vector-Project}(c, \gamma)$ produces a vector $\hat{c}$ such that with probability at least $1-\eta$, we have $|\hat{c}^\top u - c^\top u| \le \gamma$.
\end{lemma}

\begin{proof}
This is in fact a well-known corollary of Johnson-Lindenstrauss Lemma. Observe that 
$$\hat{c}^\top u = \frac{1}{k} \sum_{i=1}^k \inner{v_i,c}\inner{v_i,u}.$$
The expectation $\E[\inner{v_i,c}\inner{v_i,u}] = \E[c^\top v_iv_i^\top u] = c^\top \E[v_iv_i^\top] u = c^\top u$. The variance is bounded by $O(1/k)\le O(\gamma/\sqrt{\log n})$. Standard concentration bounds show that 
$$
\Pr[|\hat{c}^\top u - c^\top u| > \gamma/2] \le \exp(-\gamma^2k/16) \le \eta.
$$

The discretization is easy to check as with high probability the matrix $V$ with columns $v_i$'s have spectral norm at most $2\sqrt{h}$, so the vector before and after discretization can only change by $\gamma/2$.
\end{proof}

\begin{lemma}
For any number of sample $m$, there is an efficient algorithm with helper string to generate a compressed vector $\hat{c}$, such that 
$$
L(\hat{c}) \le \tilde{O}(\sqrt{1/\gamma^2 m}).
$$
\end{lemma}

\begin{proof}
We will choose $\eta = 1/m$. By Lemma~\ref{lem:vec-proj}, we know there is a compression algorithm that works with probability $1-\eta$, and has at most $O((\log 1/\eta)/\gamma^2)$ parameters. By Corollary~\ref{cor:compression}, we know
$$
L(\hat{c}) \le \tilde{O}(\eta + \sqrt{1/\gamma^2m}) \le \tilde{O}(\sqrt{1/\gamma^2 m}).
$$
\end{proof}

\subsection{Proof for Generalization Bound in \cite{neyshabur2017pac}}

We gave a compression in Lemma~\ref{lem:detcompress}, the discretization in this case is trivial just by rounding the weights to nearest multiples of $\|A\|_F/h^2$. The following lemma from \cite{neyshabur2017pac} (based on a simple induction of the noise) shows how the noises from different layers add up.

\begin{lemma}\label{lem:lipschitz}
	Let $f_A$ be a $d$-layer network with weights $A=\{A^1,\dots, A^d\}$. Then for any input $x$, weights $A$ and $\hat{A}$, if for any layer $i$, $\|A^i-\hat{A}^i\| \leq \frac{1}{d}\|A^i\|$, then we have:
	\begin{equation*}
	\|f_A(x)-f_{\hat{A}}(x)\| \leq e\|x\|\left(\prod_{i=1}^{d} \|A^i\|_2 \right) \sum_{i=1}^d \frac{\|A^i-\hat{A}^i\|_2}{\|A^i\|_2}
	\end{equation*}
\end{lemma}

Compressing each layer $i$ with $\delta=\delta = \gamma (e\|x\|d\prod_{i=1}^d \|A^i\|_2)^{-1}$ ensures $|f_A(x)-f_{\hat{A}}(x)| \leq \gamma$. Since each $\hat{A}^i$ has rank $\frac{\|A^i\|_F^2}{\delta^2\|A^i\|_2^2}$, the total number of parameters of the compressed network will be $2e^2d^2h \|x\|^2\prod_{i=1}^{d} \|A^i\|_2^2\sum_{i=1}^d\frac{\|A^i\|_F^2}{\|A^i\|_2^2}$. Therefore we can apply Theorem~\ref{thm:compress_generalize} to get the generalization bound.

%% file: fullyconnected.tex
\section{Complete Proofs for Section~\ref{sec:newmatrix}}\label{appx:newmatrix}

\subsection{Conditions}
We discussed and verified several conditions in Section~\ref{sec:stability}. Here, we formally state these conditions:
\begin{condition}\label{conditions}
	Let $S$ be the training set.
	\begin{enumerate}
		\item {\bf Layer cushion ($\mu_{i}$)}: For any layer $i$, we define the layer cushion $\mu_i$ as the largest number such that for any $x\in S$:
		$$
		\mu_{i}\|A^i\|_F\|\phi(x^{i-1})\| \leq \|A^i\phi(x^{i-1})\|
		$$
		\item {\bf Interlayer cushion ($\mu_{i,j}$)}: For any two layers $i\leq j$, we define interlayer cushion $\mu_{i,j}$ as the largest number such that for any $x\in S$:
		$$
		\mu_{i,j}\|J^{i,j}_{x^i}\|_F\|x^i\| \leq  \|J^{i,j}_{x^i}x^i\|
		$$
		Furthermore, we define minimal interlayer cushion $\icu = \min_{i\leq j\leq d} \mu_{i,j} = \min\{1/\sqrt{h^i},\min_{i< j\leq d} \mu_{i,j}\}$.
		\item {\bf Activation contraction ($c$)}: The activation contraction $c$ is defined as the smallest number such that for any layer $i$ and any $x\in S$,
		$$
		\|x^i\| \leq c \|\phi(x^i)\|
		$$
		\item {\bf Interlayer smoothness ($\rho_{\delta}$)}: Interlayer smoothness is defined the smallest number such that with probability $1-\delta$ over noise $\eta$ for any two layers $i<j$ any $x\in S$:
		$$
		\|M^{i,j}(x^{i}+\eta)-J_{x^i}^{i,j}(x^{i}+\eta)\| \leq \frac{\|\eta\|\|x^{j}\|}{\rho_{\delta}\|x^{i}\|}
		$$
	\end{enumerate}
\end{condition}

\subsection{Proofs}

\begin{proof}(of Lemma~\ref{lem:perturblayer:highprob})
	For any fixed vectors $u,v$, we have
	$$
	u^\top \hat{A}v = \frac{1}{k}\sum_{k'=1}^k u^\top Z_{k'} v = \frac{1}{k}\inner{A,M_{k'}}\inner{uv^\top, M_{k'}}.
	$$
	
	This is exactly the same as the case of Johnson-Lindenstrauss transformation. By standard concentration inequalities we know
	$$
	\Pr\left[\left|\frac{1}{k}\sum_{k'=1}^k\inner{A,M_{k'}}\inner{uv^\top, M_{k'}} - \inner{A,uv^\top}\right| \ge \epsilon\|A\|_F\|uv^\top\|_F\right] \le \exp(-k\epsilon^2).
	$$
	Therefore for the choice of $k$ we know
	$$\Pr\left[|u^\top \hat{A} v - u^\top Av\| \ge \eps\|A\|_F\|u\|\|v\|
	\right] \le \eta.$$
	
	Now for any pair of matrix/vector $(U,x)\in G$, let $u_i$ be the $i$-th row of $U$, by union bound we know with probability at least $1-\delta$ for all $u_i$ we have $|u_i^\top (\hat{A} - A) v\| \le \eps\|A\|_F\|u_i\|\|v\|$. Since $\|U(\hat{A} - A) x\|^2 = \sum_{i=1}^n (u_i^\top (\hat{A} - A)x)^2$ and $\|U\|_F^2 = \sum_{i=1}^n \|u_i\|^2$, we immediately get 
	$\|U(\hat{A} - A)x\| \ge \eps\|A\|_F\|U\|_F\|x\|$.
\end{proof}

\begin{proof}(of Lemma~\ref{lem:rotate-compress})
	We will prove this by induction. For any layer $i\geq 0$, let $\hat{x}^j_i$ be the output at layer $j$ if the weights $A^1,\dots,A^i$ in the first $i$ layers are replaced with $\tilde{A}^1,\dots,\tilde{A}^i$. The induction hypothesis is then the following:
	
	Consider any layer $i\geq 0$ and any $0<\eps\leq 1$. The following is true with probability $1-\frac{i\delta}{2d}$ over $\tilde{A}^1,\dots,\tilde{A}^i$ for any $j\geq i$:
	$$
	\|\hat{x}^j_i - x^j\|\le (i/d)\eps\|x^j\|.
	$$
	
	For the base case $i = 0$, since we are not perturbing the input, the inequality is trivial. Now assuming that the induction hypothesis is true for $i-1$, we consider what happens at layer $i$. Let $\hat{A}^i$ be the result of Algorithm~\ref{alg:matrix-proj} on $A^i$ with $\eps_i=\frac{\eps\mu_i\icu}{4cd}$ and $\eta = \frac{\delta}{6d^2h^2m}$. We can now apply Lemma~\ref{lem:perturblayer:highprob} on the set $G=\{(J^{i,j}_{x^i},x^i)|x\in S,j\geq i\}$ which has size at most $dm$.
	Let $\Delta^i = \hat{A}^i - A^i$, for any $j \ge i$ we have
	\begin{equation*}
	\|\hat{x}^j_i - x^j\| = \|(\hat{x}^j_i - \hat{x}^j_{i-1})+(\hat{x}^j_{i-1} - x^j)\| \leq \|(\hat{x}^j_i - \hat{x}^j_{i-1})\| +\|\hat{x}^j_{i-1} - x^j\|.
	\end{equation*}
	The second term can be bounded by $(i-1)\eps\|x^j\|/d$ by induction hypothesis. Therefore, in order to prove the induction, it is enough to show that the first term is bounded by $\eps/d$. We decompose the error into two error terms one of which corresponds to the error propagation through the network if activation were fixed and the other one is the error caused by change in the activations:
	\begin{align*}
	\|(\hat{x}^j_i - \hat{x}^j_{i-1})\|
	& = \|M^{i,j}(\hat{A}^i \phi(\hat{x}^{i-1})) - M^{i,j}(A^i \phi(\hat{x}^{i-1}))\|\\
	& = \|M^{i,j}(\hat{A}^i \phi(\hat{x}^{i-1})) - M^{i,j}(A^i \phi(\hat{x}^{i-1})) + J^{i,j}_{x^i}(\Delta^i \phi(\hat{x}^{i-1})) - J^{i,j}_{x^i}(\Delta^i \phi(\hat{x}^{i-1}))\| \\
	& \leq \|J^{i,j}_{x^i}(\Delta^i \phi(\hat{x}^{i-1}))\| + \|M^{i,j}(\hat{A}^i \phi(\hat{x}^{i-1})) - M^{i,j}(A^i \phi(\hat{x}^{i-1})) - J^{i,j}_{x^i}(\Delta^i \phi(\hat{x}^{i-1}))\| 
	\end{align*}
	The first term can be bounded as follows:
	\begin{align*}
	& \|J^{i,j}_{x^i}\Delta^i \phi(\hat{x}^{i-1})\|\\
	& \le (\eps\mu_i\icu/6cd)\|J^{i,j}_{x^i}\|\|A^i\|_F\|\phi(\hat{x}^{i-1})\|&& \mbox{Lemma~\ref{lem:perturblayer:highprob}}\\
	& \le (\eps\mu_i\icu/6cd)\|J^{i,j}_{x^i}\|\|A^i\|_F\|\hat{x}^{i-1}\|&& \mbox{Lipschitzness of the activation function}\\
	&\le (\eps\mu_i\icu/3cd)\|J^{i,j}_{x^i}\|\|A^i\|_F\|x^{i-1}\|&& \mbox{Induction hypothesis} \\
	&\le (\eps\mu_i\icu/3d)\|J^{i,j}_{x^i}\|\|A^i\| \|\phi(x^{i-1})\|&& \mbox{Activation Contraction} \\
	&\le (\eps\icu/3d)\|J^{i,j}_{x^i}\|\|A^i \phi(x^{i-1})\|&& \mbox{Layer Cushion} \\
	& = (\eps\icu/3d)\|J^{i,j}_{x^i}\|\|x^i\|&& x^i = A^i\phi(x^{i-1})\\
	& \le (\eps/3d) \|x^j\|&& \mbox{Interlayer Cushion} 
	\end{align*}
	The second term can be bounded as:
	\begin{align*}
	& \|M^{i,j}(\hat{A}^i \phi(\hat{x}^{i-1})) - M^{i,j}(A^i \phi(\hat{x}^{i-1})) - J^{i,j}_{x^i}(\Delta^i\phi( \hat{x}^{i-1}))\| \\
	& = \|(M^{i,j}-J^{i,j}_{x^i})(\hat{A}^i \phi(\hat{x}^{i-1})) - (M^{i,j}-J^{i,j}_{x^i})(A^i \phi(\hat{x}^{i-1}))\| \\
	& = \|(M^{i,j}-J^{i,j}_{x^i})(\hat{A}^i \phi(\hat{x}^{i-1}))\| + \|(M^{i,j}-J^{i,j}_{x^i})(A^i \phi(\hat{x}^{i-1})\|.
	\end{align*}
	Both terms can be bounded using interlayer smoothness condition of the network. First, notice that $A^i\phi(\hat{x}^{i-1}) = \hat{x}^i_{i-1}$. Therefore by induction hypothesis $\|A^i\phi(\hat{x}^{i-1}) - x^i\| \le (a-1)\eps\|x^i\|/d \le \eps\|x^i\|$. Now by interlayer smoothness property, $\|(M^{i,j}-J^{i,j}_{x^i})(A^i \phi(\hat{x}^{i-1})\| \le \frac{\|x^b\| \eps}{\rho_{\delta}} \le (\eps/3d)\|x^j\|$. On the other hand, we also know $\hat{A}^i\phi(\hat{x}^{i-1}) = \hat{x}^i_{i-1} + \Delta^i\phi(\hat{x}^{i-1})$, therefore $\|\hat{A}^i\phi(\hat{x}^{i-1}) - x^i\| \le \|A^i\phi(\hat{x}^{i-1}) - x^i\| + \|\Delta^i\phi(\hat{x}^{i-1})\| \le (i-1)\eps/d+\eps/3d \le \eps$, so again we have $\|(M^{i,j}-J^{i,j}_{x^i})(\hat{A}^i \phi(\hat{x}^{i-1}))\| \le (\eps/3d)\|x^j\|$. Putting everything together completes the induction.
\end{proof}

\begin{lemma}\label{lem:bounding_loss}
	For any fully connected network $f_A$ with $\rho_\delta\geq 3d$, any probability $0<\delta\leq 1$ and any margin $\gamma>0$, $f_A$ can be compressed (with respect to a random string) to another fully connected network $f_{\tilde{A}}$ such that for any $x\in S$, $\hat{L}_0(f_{\hat{A}}) \leq \hat{L}_{\gamma}(f_A)$ and the number of parameters in $f_{\tilde{A}}$ is at most:
	\begin{equation*}
	\tilde{O}\left(\frac{c^2d^2\max_{x\in S}\|f_A(x)\|_2^2}{\gamma^2}\sum_{i=1}^d \frac{1}{\mu_i^2\icu^2}\right)
	\end{equation*}
	where $\mu_i$, $\icu$, $c$ and $\rho_{\delta}$ are layer cushion, interlayer cushion, activation contraction and interlayer smoothness defined in Definitions~\ref{def:layercushion},\ref{def:interlayercushion},\ref{def:activationcontraction} and \ref{def:interlayersmoothness} respectively.
\end{lemma}
\begin{proof}(of Lemma~\ref{lem:bounding_loss})
	If $\gamma^2 > 2\max_{x\in S}\|f_A(x)\|_2^2$, for any pair $(x,y)$ in the training set we have $|f_A(x)[y]  - \max_{i\neq y} f_A(x)[j]|^2 \leq 2\max_{x\in S}\|f_A(x)\|_2^2 \leq \gamma$ which means the output margin cannot be greater than $\gamma$ and therefore $\hat{L}_{\gamma}(f_A)=1$ which proves the statement. If $\gamma^2\leq 2\max_{x\in S}\|f_A(x)\|_2^2$, by setting $\eps^2 = \gamma^2/2\max_{x\in S}\|f_A(x)\|_2^2$ in Lemma~\ref{lem:rotate-compress}, we know that for any $x\in S$, $\|f_A(x) - f_{\tilde{A}}(x)\|_{2}\le \gamma/\sqrt{2}$. For any $(x,y)$, if the margin loss on the right hand side is one then the inequality holds. Otherwise, the output margin in $f_{\tilde{A}}$ is greater than $\gamma$ which means in order for classification loss of $f_{A}$ to be one, we neet to have $\|f_A(x) - f_{\tilde{A}}(x)\|_{2}> \gamma/\sqrt{2}$ which is not possible and that completes the proof.
\end{proof}

\begin{proof}(of Theorem~\ref{thm:fullyconnected_generalization})
	We show the generalization by bounding the covering number of the network with weights $\tilde{A}$. We already demonstrated that the original network with weights $A$ can be approximated with another network with weights $\tilde{A}$ and less number of parameters. In order to get a covering number, we need to find out the required accuracy for each parameter in the second network to cover the original network. We start by bounding the norm of the weights $\tilde{A}^i$.
	
	Because of positive homogeneity of ReLU activations, we can assume without loss of generality that the network is balanced, i.e for any $i\neq j$, $\|A_i\|_F = \|A_j\|_F=\beta$ (otherwise, one could rebalance the network before approximation and cushion in invariant to this rebalancing). Therefore, for any $x\in S$ we have:
	\begin{align*}
	\beta^d = \prod_{i=1}^d\|A^i\| \leq \frac{c\|x^1\|}{\|x\| \mu_1}\prod_{i=2}^d\|A^i\|\leq \frac{c^2\|x^2\|}{\|x\| \mu_1\mu_2}\prod_{i=2}^d\|A^i\|\leq \frac{c^{d} \|f_A(x)\|}{\|x\|\prod_{i=1}^d \mu_i}
	\end{align*}
	By Lemma~\ref{lem:rotate-compress}, $\|\tilde{A^i}\|_F \leq \beta(1+1/d)$. We know that $\tilde{A}^i = \frac{1}{k}\sum_{k'=1}^k \langle A^i, M_{k'}\rangle M_{k'}$ where $\langle A^i, M_{k'}\rangle$ are the parameters. Therefore, if $\hat{A}^i$ correspond to the weights after approximating each parameter in $\tilde{A}^i$ with accuracy $\nu$, we have:
	$\|\hat{A}^i-\tilde{A}^i\|_F \leq \sqrt{k}h\nu \leq \sqrt{q}h\nu$ where $q$ is the total number of parameters. Now by Lemma \ref{lem:lipschitz}, we get:
	
	\begin{align*}
	|\ell_\gamma(f_{\tilde{A}}(x),y) -\ell_\gamma(f_{\hat{A}}(x),y)| &\leq
	\frac{2e}{\gamma}\|x\|\left(\prod_{i=1}^{d} \|\tilde{A}^i\| \right) \sum_{i=1}^d \frac{\|\tilde{A}^i-\hat{A}^i\|}{\|\tilde{A}^i\|} < \frac{e^2}{\gamma}\|x\| \beta^{d-1}\sum_{i=1}^d \|\tilde{A}^i-\hat{A}^i\|_F\\
	&\leq \frac{e^2c^d \|f_A(x)\| \sum_{i=1}^d \|\tilde{A}^i-\hat{A}^i\|_F}{\gamma\beta\prod_{i=1}^d \mu_i} \leq \frac{ qh\nu}{\beta}\\
	\end{align*}
	where the last inequality is because by Lemma~\ref{lem:bounding_loss}, $\frac{e^2d \|f_A(x)\|}{\gamma\beta\prod_{i=1}^d \mu_i } < \sqrt{q}$.
	Since the absolute value of each parameter in layer $i$ is at most $\beta h$, the logarithm of number of choices for each parameter in order to get $\eps$-cover is $\log(qh^2/\eps) \leq 2\log(qh/\eps)$ which results in the covering number $2q\log(kh/\eps)$. Bounding the Rademacher complexity by Dudley entropy integral completes the proof.
\end{proof}

%% file: convolution.tex
\section{Convolutional Neural Networks}
\renewcommand{\S}{\mathcal{S}}
\label{sec:convolution:appendix}
In this section we give a compression algorithm for convolutional neural networks, and prove Theorem~\ref{thm:convolution:main}.

We start by developing some notations to work with convolutions and product of tensors. For simplicity of notation, for any $k'\leq k$, we define a product operator $\times_{k'}$ that given a $k$th-order tensor $Y$ and a $k'$ order  tensor $Z$ with a matching dimensionality to the last $k'$-dimensions of $Y$, vectorizes the last $k'$ dimensions of each tensor and returns a $k-k'$th order tensor as follows:
$$
(Y \times_{k'} Z)_{i_1,\dots,i_{k-k'}} = \langle Y_{i_1,\dots,i_{k-k'}},Z\rangle=\langle \text{vec}(Y_{i_1,\dots,i_{k-k'}}),\text{vec}(Z)\rangle
$$

Let $X\in \R^{h \times n_1\times n_2}$ be an $n\times n$ image where $h$ is the number of features for each pixel. We denote the $\kappa\times \kappa$ sub-image of $X$ starting from pixel $(i,j)$ by $X_{(i,j),\kappa}\in\R^{h\times \kappa \times \kappa}$. Let $A\in \R^{h' \times h \times \kappa \times \kappa}$ be a convolutional weight tensor. Now the convolution operator with stride $s$ can be defined as follows:
$$
(A*_{s}X)_{i,j} =  A \times_3 X_{(s(i-1)+1,s(j-1)+1),\kappa} \qquad \forall 1\leq i \leq \lfloor \frac{n_1 - \kappa}{s}\rfloor,  1\leq i \leq \lfloor \frac{n_2 - \kappa}{s}\rfloor
$$
where $n'_1 = \lfloor \frac{n_1 - \kappa}{s}\rfloor$, $n'_2=\lfloor \frac{n_2 - \kappa}{s}\rfloor$ and $A*_{s}X \in \R^{h'\times n'_1\ \times n'_2}$.

As we discussed in Section~\ref{sec:convolution}, we will actually have a different set of weights at each convolution location. Let $\hat{A}_{(i,j)}\in \R^{h'\times h\times \kappa\times\kappa} (i\in[n'_1],j\in[n'_2])$ be a set of weights for each location, we use the notation $\hat{A}*_{s}X$ to denote
$$
((\hat{A}*_{s}X)_{i,j}) = \hat{A}_{(i,j)} \times_3 X_{(s(i-1)+1,s(j-1)+1),\kappa} \qquad \forall 1\leq i \leq \lfloor \frac{n_1 - \kappa}{s}\rfloor,  1\leq i \leq \lfloor \frac{n_2 - \kappa}{s}\rfloor.
$$
The $\hat{A}_{(i,j)}$'s will be generated by Algorithm~\ref{alg:matrix-proj-p} and are $p$-wise independent.

Let $\kappa_i$ be the filter size and $s_i$ be the stride in layer $i$ of the convolutional network. Then for any $i>1$, $x^{i+1} = \phi(A^i *_{s_i} x^i)$. Furthermore, since the activation functions are ReLU, we have $x^j = M^{ij}(x^i)=J^{ij}_{x^i} \times_3 x^i$.

In the rest of this section, we will first describe the compression algorithm Matrix-Project-Conv (Algorithm~\ref{alg:matrix-proj-p}) and show that the output of this algorithm behaves similar to Gaussian noise (similar to Lemma~\ref{lem:perturblayer:highprob}). Then we will follow the same strategy as the feed-forward case and give the full proof.

\subsection{$p$-wise Independent Compression}

\begin{algorithm}
	\caption{Matrix-Project-Conv($A$, $\eps$, $\eta$, $n'_1\times n'_2$)}
	\begin{algorithmic}
		\REQUIRE Convolution Tensor $A\in \R^{h' \times h \times \kappa \times \kappa}$, error parameter $\eps$, $\eta$.
		\ENSURE 
Generate $n'_1\times n'_2$ different tensors $\hat{A}_{(i,j)}((i,j)\in [n'_1]\times [n'_2])$ that satisfies Lemma~\ref{lem:conv-perturblayer:highprob}
	%
\STATE Let $k = \frac{Q\lceil\kappa/s\rceil^2\log^2 1/\eta}{\epsilon^2}$ for a large enough universal constant $Q$.
\STATE Let $p = \log(1/\eta)$
\STATE
Sample a uniformly random subspace $\S$ of $h'\times h\times \kappa\times \kappa$ of dimension $k\times p$
\FOR{each $(i,j)\in [n'_1]\times [n'_2]$}		
		\STATE Sample $k$ matrices $M_1, M_2,...,M_k \in \N(0,1)^{h' \times h \times \kappa \times \kappa}$ with random i.i.d. entries.
		\FOR{$k'=1$ to $k$}
		\STATE Let $M'_{k'} = \sqrt{hh'\kappa^2/kp}\cdot \mbox{Proj}_\S(M_{k'})$.
		\STATE Let $Z_{k'} = \inner{A,M'_{k'}} M'_{k'}$.
		\ENDFOR
		\STATE Let $\hat{A}_{(i,j)} =\frac{1}{k}\sum_{k'=1}^k Z_{k'}$
		\ENDFOR
	\end{algorithmic}
	\label{alg:matrix-proj-p}
\end{algorithm}

The weights in convolutional neural networks have inherent correlation due to the architecture, as the weights are shared across different locations. However, in order to randomly compress the weight tensors, we need to break this correlation and try to introduce independent perturbations at every location. The procedure is described as Algorithm~\ref{alg:matrix-proj-p}.

The goal of Algorithm~\ref{alg:matrix-proj-p} is to generate different compressed filters $\hat{A}_{i,j}$ such that the total number of parameters is small, and at the same time $\hat{A}_{i,j}$'s behave very similarly to applying Algorithm~\ref{alg:matrix-proj} $A$ for each location independently. We formalize these two properties in the following two lemmas:

\begin{lemma}\label{lem:pwiseparametercount}
Given a helper string that contains all of the $M'$ matrices used in Algorithm~\ref{alg:matrix-proj-p}, then it is possible to compute all of $\hat{A}_{(i,j)}$'s based on $\mbox{Proj}_\S(A)$. Since $\S$ is a $kp$ dimensional subspace $\mbox{Proj}_\S(A)$ has $kp$ parameters.
\end{lemma}

\begin{proof}
By Algorithm~\ref{alg:matrix-proj-p} we know $\hat{A}_{(i,j)}$'s are average of the $Z$ matrices, and $Z_{k'} = \inner{A,M'_{k'}} M'_{k'}$. Since $M'_{k'}\in \mathcal{S}$, we know $\inner{A,M'_{k'}} = \inner{\mbox{Proj}_\S(A), M'_{k'}}$. Hence $Z_{k'} = \inner{\mbox{Proj}_\S(A), M'_{k'}} M'_{k'}$ only depends on $\mbox{Proj}_\S(A)$ and $M'_{k'}$. 
\end{proof}

\begin{lemma}\label{lem:pwiseindep}
The random matrices $\hat{A}_{(i,j)}$'s generated by Algorithm~\ref{alg:matrix-proj-p} are $p$-wise independent. Moreover, for any $\hat{A}_{(i,j)}$, the marginal distribution of the $M'$ matrices are i.i.d. Gaussian with variance 1 in every direction.
\end{lemma}

\begin{proof}
Take any subset of $p$ random matrices $\hat{A}_{(i_1,j_1)},...,\hat{A}_{(i_p,j_p)}$ generated by Algorithm~\ref{alg:matrix-proj-p}. 
We are going to consider the joint distribution of all the $M'$ matrices used in generating these $\hat{A}$'s ($k\times p$ of them) and the subspace $\S$.

Consider the following procedure: generate $k\times p$ random matrices $M'_1,M'_2...,M'_{kp}$ from $N(0,1)^{h'\times h\times \kappa\times \kappa}$, and let $\S$ be the span of these $kp$ vectors. By symmetry of Gaussian vectors, we know $\S$ is a uniform random subspace of dimension $kp$. 

Now we sample from the same distribution in a different order: first sample a uniform random subspace $\S$ of dimension $kp$, then sample $kp$ random Gaussian matrices within this subspace (which can be done by sample a Gaussian in the entire space and then project to this subspace). This is exactly the procedure described in Algorithm~\ref{alg:matrix-proj-p}.

Therefore, the $M'$ matrices used in generating these $\hat{A}$'s are independent, as a result the $\hat{A}_{(i,j)}$'s are also independent. The equivalence also shows that the marginal distributions of $M'$ are i.i.d. spherical Gaussians.
(Note that the reason this is limited to $p$-wise independence is that if we look at more than $kp$ random matrices from the subspace $\S$, they do not have the same distribution as Gaussian random matrices; the latter would span a subspace of dimension higher than $kp$.)
\end{proof}

Although the $\hat{A}_{(i,j)}$'s are only $p$-wise independent, when $p = \log 1/\eta$ we can show that they behave similarly to fully independent random filters. We defer the technical concentration bounds to the end of this section (Section~\ref{subsec:concentration}).

Using this compression, we will prove that the noise generated at each layer behaves similar to a random vector. In particular it does not correlate with any fixed tensor, as long as the norms of the tensor is {\em well-distributed}:

\begin{definition}
Let $U\in\R^{h'\times n_1'\times n_2'\times n_u}$, we say $U$ is $\beta$ well-distributed if for any $i,j\in [n_1']\times [n'_2]$, $\|U_{:,j,k,:}\|_F \le \frac{\beta}{\sqrt{n'_1 n'_2}}\|U\|_F$. 
\end{definition}

Intuitively, $U$ is well-distributed if no spacial location of $U$ has a norm that is significantly larger than the average. Now we are ready to show the noise generated by this procedure behaves very similar to a random Gaussian (this is a generalization of Lemma~\ref{lem:perturblayer:highprob}):

\begin{lemma}\label{lem:conv-perturblayer:highprob}
For any $0<\delta,\eps\leq 1$, et $G=\{(U^i,V^i)\}_{i=1}^m$ be a set of matrix/vector pairs of size $m$ where $U\in\R^{h'\times n'_1\times n'_2\times n_u}$\footnote{$U$ can have more than $4$-orders, here we vectorize all the remaining directions in $U$ as it does not change the proof.} and $V\in\R^{h\times n_1\times n_2}$, let $\hat{A}_{(i,j)}\in \R^{h\times h'}$ be the output of Algorithm~\ref{alg:matrix-proj-p} with $\eta = \delta/n$ and $\Delta_{(i,j)} = \hat{A}_{(i,j)} - A$. Suppose all of $U$'s are $\beta$-well-distributed. With probability at least $1-\delta$ we have for any $(U,V)\in G$, $\|U\times_3 (\Delta*_s V)\| \leq \frac{\eps\beta}{\sqrt{n'_1n'_2}}\|A\|_F\|U\|_F\|V\|_F$.
\end{lemma}

\begin{proof}
We will first expand out $U\times_3 (\Delta*_s V)$:
$$
U\times_3 (\Delta*_s V) = \sum_{i=1}^{n_1'}\sum_{j=1}^{n_2'} (U_{:,i,j,:}\otimes V_{(s(i-1)+1,s(j-1)+1), \kappa})\times_4 (\hat{A}_{(i,j)} - A).
$$
In this expression, $(U_{:,i,j,:}\otimes V_{(s(i-1)+1,s(j-1)+1), \kappa})$ generates a 5-th order tensor (2 from $U$ and 3 from $V$), the order of dimensions is that $V$ takes coordinates number 3,4,5 (with dimensions $h\times\kappa\times\kappa$), the first dimension of $U$ takes the 2nd coordinate and the 4-th dimension of $U$ takes the 1st coordinate. The result of $(U_{:,i,j,:}\otimes V_{(s(i-1)+1,s(j-1)+1), \kappa})\times_4 (\hat{A}_{(i,j)} - A)$ is a vector of dimension $n_u$ (because the first 4 dimensions are removed in the inner-product). 

Now let us look at the terms in this sum, let $X_{i,j} = (U_{:,i,j,:}\otimes V_{(s(i-1)+1,s(j-1)+1), \kappa})\times_4 \hat{A}_{(i,j)}$. Let $M'_1,...,M'_k$ be the random matrices used when computing $\hat{A}_{(i,j)}$ (for simplicity we omit the indices for $i,j$), then we have
$$
X_{i,j} = \frac{1}{k}\sum_{l = 1}^k [(U_{:,i,j,:}\otimes V_{(s(i-1)+1,s(j-1)+1), \kappa})\times_4 M'_l] \inner{A,M'_l}. 
$$

Since the marginal distribution of $M'_l$ is a spherical Gaussian, it's easy to check that $\E[X_{i,j}] = (U_{:,i,j,:}\otimes V_{(s(i-1)+1,s(j-1)+1), \kappa})\times_4 A$. Also, the first term $[(U_{:,i,j,:}\otimes V_{(s(i-1)+1,s(j-1)+1), \kappa})\times_4 M'_l]$ is a Gaussian random vector whose expected squared norm is $\|U_{:,i,j,:}\|_F^2\|V_{(s(i-1)+1,s(j-1)+1), \kappa}\|_F^2$; the second term $\inner{A,M'_l}$ is a Gaussian random variable with variance $\|A\|_F^2$. By the relationship between Gaussians and subexponential random variables, there exists a universal constant $Q'$ such that $[(U_{:,i,j,:}\otimes V_{(s(i-1)+1,s(j-1)+1), \kappa})\times_4 M'_l] \inner{A,M'_l}$ is a vector whose norm is $Q'\|U_{:,i,j,:}\|_F\|V_{(s(i-1)+1,s(j-1)+1), \kappa}\|_F\|A\|_F$-subexponential. The average of $k$ independent copies lead to a random vector $X_{i,j}$ whose norm is $\sigma_{i,j}$-subexponential, where $\sigma_{i,j} = \frac{Q'}{\sqrt{k}}\|U_{:,i,j,:}\|_F\|V_{(s(i-1)+1,s(j-1)+1), \kappa}\|_F\|A\|_F$\footnote{Notice that here this average over $k$ independent copies actually has a better tail than a subexponential random variable. However for simplicity we are not trying to optimize the dependencies on $\log$ factors here.}.

By Lemma~\ref{lem:pwiseindep} we know $X_{i,j}$'s are $p$-wise independent. Now we can apply Corollary~\ref{cor:pwiseconcentration:vector} to the sum of $X_{i,j}$'s. Let $\sigma = \sqrt{\sum_{i=1}^{n'_1}\sum_{j=1}^{n'_2} \sigma_{i,j}^2}$, then we know
$$
\Pr[\|U\times_3 (\Delta*_s V)\| \ge 12\sigma p] \le 2^{-p} = \eta = \delta/m. 
$$
Union bound over all $(U,V)$ pairs, we know with probability at least $1-\delta$, we have $\|U\times_3 (\Delta*_s V)\| \le 12\sigma p$ for all $(U,V)$. 

Finally, we will try to relate $12\sigma p$ with $\frac{\eps\beta}{\sqrt{n'_1n'_2}}\|A\|_F\|U\|_F\|V\|_F$.

\begin{align*}
\sigma & = \sqrt{\sum_{i=1}^{n'_1}\sum_{j=1}^{n'_2} \sigma_{i,j}^2}\\
& = \sqrt{\sum_{i=1}^{n'_1}\sum_{j=1}^{n'_2}\frac{(Q')^2}{k}\|U_{:,i,j,:}\|_F^2\|V_{(s(i-1)+1,s(j-1)+1), \kappa}\|_F^2\|A\|_F^2} \\
& = \frac{Q'}{\sqrt{k}} \|A\|_F\sqrt{\sum_{i=1}^{n'_1}\sum_{j=1}^{n'_2}\|U_{:,i,j,:}\|_F^2\|V_{(s(i-1)+1,s(j-1)+1), \kappa}\|_F^2} \\
& \le \frac{Q'\beta}{\sqrt{n'_1n'_2}\sqrt{k}} \|A\|_F\|U\|_F\sqrt{\sum_{i=1}^{n'_1}\sum_{j=1}^{n'_2}\|V_{(s(i-1)+1,s(j-1)+1), \kappa}\|_F^2} \\
& \le \frac{Q'\beta\lceil\kappa/s\rceil}{\sqrt{n'_1n'_2}\sqrt{k}}\|A\|_F\|U\|_F\|V\|_F.
\end{align*}
Here the first inequality is by the assumption that all $U$'s are $\beta$-well-distributed. The second inequality is true because each entry in $V$ appears in at most $\lceil\kappa/s\rceil^2$ entries of $V_{(s(i-1)+1,s(j-1)+1), \kappa}$. Therefore, when $k$ is set to $144(Q')^2\lceil\kappa/s\rceil^2p^2/\epsilon^2 = O(\frac{\lceil\kappa/s\rceil^2\log^2 1/\eta}{\epsilon^2})$, we have $12\sigma p \le \frac{\eps\beta}{\sqrt{n'_1n'_2}}\|A\|_F\|U\|_F\|V\|_F$ as desired.
\end{proof}

\subsection{Generalization Bounds for Convolutional Neural Networks}

Next we will use Algorithm~\ref{alg:matrix-proj-p} to compress the neural network and prove generalization bounds. 
Similar to the feed-forward case, our first step is to show bound the perturbation of the output based on the noise introduced at each layer. This is captured by the following lemma (generalization of Lemma~\ref{lem:rotate-compress})

\begin{lemma}\label{lem:conv-compress}
	For any convolutional neural network $f_A$ with $\rho_\delta\geq 3d$, any probability $0<\delta\leq 1$ and any error $0<\eps\leq 1$, Algorithm~\ref{alg:matrix-proj-p} generates weights $\tilde{A}^i_{(a,b)}$ for each layer $i$ and each convolution location $(a,b)$  with $\tilde{O}\left(\frac{c^2d^2\beta^2}{\eps^2}\cdot \sum_{i=1}^d \frac{\lceil\kappa_i/s_i\rceil^2}{\mu_i^2\icu^2}\right)$ total parameters such that with probability $1-\delta/2$ over the generated weights $\tilde{A}_{(i,j)}$, for any $x\in S$:
	$$
	\|f_A(x) - f_{\tilde{A}}(x)\| \le \eps\|f_A(x)\|.
	$$
	where $\mu_i$, $\icu$, $c$, $\rho_{\delta}$ and $\beta$ are layer cushion, interlayer cushion, activation contraction, interlayer smoothness and well-distributedness of Jacobian defined in Definitions~\ref{def:layercushion},\ref{def:interlayercushionLconv},\ref{def:activationcontraction}, \ref{def:interlayersmoothness} and \ref{def:welldistributed:jacobian} respectively.
\end{lemma}

\begin{proof}
	We will prove this by induction. For any layer $i\geq 0$, let $\hat{x}^j_i$ be the output at layer $j$ if the weights $A^1,\dots,A^i$ in the first $i$ layers are replaced with $\{\tilde{A}^1_{(a,b)}\},\dots,\{\tilde{A}^i_{(a,b)}\}$. The induction hypothesis is then the following:
	
	Consider any layer $i\geq 0$ and any $0<\eps\leq 1$. The following is true with probability $1-\frac{i\delta}{2d}$ over $\tilde{A}^1,\dots,\tilde{A}^i$ for any $j\geq i$:
	$$
	\|\hat{x}^j_i - x^j\|\le (i/d)\eps\|x^j\|.
	$$
	
	(Note that although $x$ is now a 3-tensor, we still use $\|x\|$ to denote $\|x\|_F$ as we never use any other norm of $x$.)
	
	For the base case $i = 0$, since we are not perturbing the input, the inequality is trivial. Now assuming that the induction hypothesis is true for $i-1$, we consider what happens at layer $i$. Let $\tilde{A}^i$ be the result of Algorithm~\ref{alg:matrix-proj} on $A^i$ with $\eps_i=\frac{\eps\mu_i\icu}{4cd\beta}$ and $\eta = \frac{\delta}{6d^2h^2m}$. We can now apply Lemma~\ref{lem:perturblayer:highprob} on the set $G=\{(J^{i,j}_{x^i},x^i)|x\in S,j\geq i\}$ which has size at most $dm$.
	Let $\Delta^i_{(a,b)} = \tilde{A}^i_{(a,b)} - A^i$ ($(a,b)\in [n^i_1]\times[n^i_2])$, for any $j \ge i$ we have
	\begin{equation*}
	\|\hat{x}^j_i - x^j\| = \|(\hat{x}^j_i - \hat{x}^j_{i-1})+(\hat{x}^j_{i-1} - x^j)\| \leq \|(\hat{x}^j_i - \hat{x}^j_{i-1})\| +\|\hat{x}^j_{i-1} - x^j\|.
	\end{equation*}
	The second term can be bounded by $(i-1)\eps\|x^j\|/d$ by induction hypothesis. Therefore, in order to prove the induction, it is enough to show that the first term is bounded by $\eps/d$. We decompose the error into two error terms one of which corresponds to the error propagation through the network if activation were fixed and the other one is the error caused by change in the activations:
	\begin{align*}
	\|(\hat{x}^j_i - \hat{x}^j_{i-1})\|
	& = \|M^{i,j}(\tilde{A}^i *_s\phi(\hat{x}^{i-1})) - M^{i,j}(A^i *_s\phi(\hat{x}^{i-1}))\|\\
	& = \|M^{i,j}(\tilde{A}^i*_s \phi(\hat{x}^{i-1})) - M^{i,j}(A^i *
_s\phi(\hat{x}^{i-1})) + J^{i,j}_{x^i}\times_3(\Delta^i *_s\phi(\hat{x}^{i-1})) - J^{i,j}_{x^i}\times_3(\Delta^i *_s\phi(\hat{x}^{i-1}))\| \\
	& \leq \|J^{i,j}_{x^i}\times_3(\Delta^i *_s\phi(\hat{x}^{i-1}))\| + \|M^{i,j}(\tilde{A}^i*_s \phi(\hat{x}^{i-1})) - M^{i,j}(A^i *_s\phi(\hat{x}^{i-1})) - J^{i,j}_{x^i}\times_3(\Delta^i *_s\phi(\hat{x}^{i-1}))\| 
	\end{align*}
	The first term can be bounded as follows:
	\begin{align*}
	& \|J^{i,j}_{x^i}\times_3(\Delta^i *_s\phi(\hat{x}^{i-1}))\|\\
	& \le (\eps\mu_i\icu/6cd)\cdot \frac{1}{\sqrt{n^i_1n^i_2}}\|J^{i,j}_{x^i}\|_F\|A^i\|_F\|\phi(\hat{x}^{i-1})\|&& \mbox{Lemma~\ref{lem:conv-perturblayer:highprob}}\\
	& \le (\eps\mu_i\icu/6cd)\cdot \frac{1}{\sqrt{n^i_1n^i_2}}\|J^{i,j}_{x^i}\|_F\|A^i\|_F\|\hat{x}^{i-1}\|&& \mbox{Lipschitzness of the activation function}\\
	&\le (\eps\mu_i\icu/3cd)\cdot \frac{1}{\sqrt{n^i_1n^i_2}}\|J^{i,j}_{x^i}\|_F\|A^i\|_F\|x^{i-1}\|&& \mbox{Induction hypothesis} \\
	&\le (\eps\mu_i\icu/3d)\cdot \frac{1}{\sqrt{n^i_1n^i_2}}\|J^{i,j}_{x^i}\|_F\|A^i\| \|\phi(x^{i-1})\|&& \mbox{Activation Contraction} \\
		&\le (\eps\icu/3d)\cdot \frac{1}{\sqrt{n^i_1n^i_2}}\|J^{i,j}_{x^i}\|_F\|A^i *_s\phi(x^{i-1})\|&& \mbox{Layer Cushion} \\
	& = (\eps\icu/3d)\cdot \frac{1}{\sqrt{n^i_1n^i_2}}\|J^{i,j}_{x^i}\|_F\|x^i\|&& x^i = A^i*_s\phi(x^{i-1})\\
	& \le (\eps/3d) \|x^j\|&& \mbox{Interlayer Cushion}
	%
	%
	\end{align*}
The second term can be bounded as:
\begin{align*}
& \|M^{i,j}(\tilde{A}^i*_s \phi(\hat{x}^{i-1})) - M^{i,j}(A^i*_s \phi(\hat{x}^{i-1})) - J^{i,j}_{x^i}\times_3(\Delta^i*_s\phi( \hat{x}^{i-1}))\| \\
& = \|(M^{i,j}-J^{i,j}_{x^i})\times_3(\tilde{A}^i*_s \phi(\hat{x}^{i-1})) - (M^{i,j}-J^{i,j}_{x^i})\times_3(A^i*_s\phi(\hat{x}^{i-1}))\| \\
& = \|(M^{i,j}-J^{i,j}_{x^i})\times_3(\tilde{A}^i *_s\phi(\hat{x}^{i-1}))\| + \|(M^{i,j}-J^{i,j}_{x^i})\times_3(A^i*_s \phi(\hat{x}^{i-1})\|.
\end{align*}
Both terms can be bounded using interlayer smoothness condition of the network. First, notice that $A^i*_s\phi(\hat{x}^{i-1}) = \hat{x}^i_{i-1}$. Therefore by induction hypothesis $\|A^i*_s\phi(\hat{x}^{i-1}) - x^i\| \le (i-1)\eps\|x^i\|/d \le \eps\|x^i\|$. Now by interlayer smoothness property, $\|(M^{i,j}-J^{i,j}_{x^i})\times_3(A^i*_s \phi(\hat{x}^{i-1})\| \le \frac{\|x^j\| \eps}{\rho_{\delta}} \le (\eps/3d)\|x^j\|$. On the other hand, we also know $\tilde{A}^i*_s\phi(\hat{x}^{i-1}) = \hat{x}^i_{i-1} + \Delta^i*_s\phi(\hat{x}^{i-1})$, therefore $\|\tilde{A}^i*_s\phi(\hat{x}^{i-1}) - x^i\| \le \|A^i*_s\phi(\hat{x}^{i-1}) - x^i\| + \|\Delta^i*_s\phi(\hat{x}^{i-1})\| \le (i-1)\eps/d+\eps/3d \le \eps$, so again we have $\|(M^{i,j}-J^{i,j}_{x^i})\times_3(\tilde{A}^i*_s \phi(\hat{x}^{i-1}))\| \le (\eps/3d)\|x^j\|$. Putting everything together completes the induction.
\end{proof}

Now we are ready to prove Theorem~\ref{thm:convolution:main}

\begin{proof}
We show the generalization by bounding the covering number of the network with weights $\tilde{A}$. We already demonstrated that the original network with weights $A$ can be approximated with another network with weights $\tilde{A}$ and less number of parameters. In order to get a covering number, we need to find out the required accuracy for each parameter in the second network to cover the original network. We start by bounding the norm of the weights $\tilde{A}^i$.

Because of positive homogeneity of ReLU activations, we can assume without loss of generality that the network is balanced, i.e for any $i\neq j$, $\|A_i\|_F = \|A_j\|_F=\tau$ (otherwise, one could rebalance the network before approximation and cushion in invariant to this rebalancing). Therefore, for any $x\in S$ we have:
\begin{align*}
\tau^d = \prod_{i=1}^d\|A^i\|_F \leq \frac{c\|x^1\|}{\|x\| \mu_1}\prod_{i=2}^d\|A^i\|_F\leq \frac{c^2\|x^2\|}{\|x\| \mu_1\mu_2}\prod_{i=2}^d\|A^i\|_F\leq \frac{c^{d} \|f_A(x)\|}{\|x\|\prod_{i=1}^d \mu_i}
\end{align*}
By Lemma~\ref{lem:conv-compress} and Lemma~\ref{lem:pwiseparametercount}, we know $\mbox{Proj}_\mathcal{S} A^i$ are the parameter. 
 Therefore, if $\hat{A}^i$ correspond to the weights after approximating each parameter in $\tilde{A}^i$ with accuracy $\nu$, we have:
$\|\hat{A}^i-\tilde{A}^i\|_F \leq \sqrt{k}h\nu \leq \sqrt{q}h\nu$ where $q$ is the total number of parameters. Now by Lemma \ref{lem:lipschitz}, we get:

\begin{align*}
	|\ell_\gamma(f_{\tilde{A}}(x),y) -\ell_\gamma(f_{\hat{A}}(x),y)| &\leq
	\frac{2e}{\gamma}\|x\|\left(\prod_{i=1}^{d} \|\tilde{A}^i\| \right) \sum_{i=1}^d \frac{\|\tilde{A}^i-\hat{A}^i\|}{\|\tilde{A}^i\|} < \frac{e^2}{\gamma}\|x\| \tau^{d-1}\sum_{i=1}^d \|\tilde{A}^i-\hat{A}^i\|_F\\
	&\leq \frac{e^2c^d \|f_A(x)\| \sum_{i=1}^d \|\tilde{A}^i-\hat{A}^i\|_F}{\gamma\tau\prod_{i=1}^d \mu_i} \leq \frac{ qh\nu}{\tau}\\
\end{align*}
where the last inequality is because by Lemma~\ref{lem:bounding_loss}, $\frac{e^2d \|f_A(x)\|}{\gamma\tau\prod_{i=1}^d \mu_i } < \sqrt{q}$.
Since the absolute value of each parameter in layer $i$ is at most $\tau h$, the logarithm of number of choices for each parameter in order to get $\eps$-cover is $\log(qh^2/\eps) \leq 2\log(qh/\eps)$ which results in the covering number $2q\log(kh/\eps)$. Bounding the Rademacher complexity by Dudley entropy integral completes the proof.
\end{proof}

Similar to the discussions at the end of Section~\ref{sec:newmatrix}, we can use distance to initialization and remove outliers. More concretely, we can get the following corollary

\begin{corollary}
	For any convolutional neural network $f_A$ with $\rho_\delta\geq 3d$,any probability $0<\delta\leq 1$ and any margin $\gamma$, Algorithm~\ref{alg:matrix-proj-p} generates weights $\tilde{A}$ for the network $f_{\tilde{A}}$ such that with probability $1-\delta$ over the training set and $f_{\tilde{A}}$:
$$
L_0(f_{\tilde{A}}) \leq \hat{L}_\gamma(f_A) + \zeta+ \tilde{O}\left(\sqrt{\frac{c^2d^2 \max_{x\in S}\|f_A(x)\|_2^2 \sum_{i=1}^d \frac{\beta^2(\lceil\kappa_i/s_i\rceil)^2}{\mu_i^2\icu^2}}{\gamma^2 m}}\right)
$$
where $\mu_i$, $\icu$, $c$ and $\rho_{\delta}$ are layer cushion, interlayer cushion, activation contraction and interlayer smoothness defined in Definitions~\ref{def:layercushion},\ref{def:interlayercushionLconv},\ref{def:activationcontraction} and \ref{def:interlayersmoothness} respectively and measured on a $1-\zeta$ fraction of the training set $S$.
\end{corollary}

\subsection{Concentration Inequalities for Sum of $p$-wise Independent Variables}
\label{subsec:concentration}
In this section we prove a technical lemma that shows the sum of $p$-wise independent subexponential random variables have strong concentration properties. Previously similar results were known for Bernoulli random variables \cite{pelekis2015hoeffding}, the approach we take here is very similar.

\begin{definition}
A random variable $X$ is $\sigma$-subexponential if for all $k > 0$, $\E[|X-\E[X]|^k] \le \sigma^k k^k$.
\end{definition}

The following lemma will imply concentration

\begin{lemma}\label{lem:pwiseconcentration}
Let $X_1,X_2,...,X_n$ be random variables where $X_i$ is $\sigma_i$-subexponential. Let $\sigma^2 = \sum_{i=1}^n \sigma_i^2$, $X = \sum_{i=1}^n X_i$. If $X_i$'s are $p$-wise independent 
$$
\E[(X-\E[X])^p] \le (3\sigma)^p\cdot (2p)^p.
$$
In particular, for all $t > 1$,
$$
\Pr[|X-\E[X]| \ge 6\sigma p t] \le 1/t^p.
$$
\end{lemma}

\begin{proof}
Let $Y_i = X_i - \E[X_i]$ and $Y = X-\E[X]$, we will compute $\E[Y^p]$. 

$$
\E[Y^p] = \sum_{a, a_i\in \N, \sum a_i = p} \frac{p!}{\prod_{i=1}^n a_i!}\E[\prod_{i=1}^n Y_i^{a_i}] = \sum_{a, a_i\in \mathbb{N}, \sum a_i = p} \frac{p!}{\prod_{i=1}^n a_i!}\prod_{i=1}^n \E[Y_i^{a_i}]
$$

Here the last step is because $Y_i$'s are $p$-wise independent. Now, notice that $\E[Y_i] = 0$. Therefore, as long as one of the $a_i$'s is equal to 1, we have $\prod_{i=1}^n \E[Y_i^{a_i}] = 0$. All the remaining terms are terms with $a_i$'s either equal to 0 or at least 2. Let $\mathcal{A}$ be the set of such $a$'s, then we have
$$
\E[Y^p] = \sum_{a\in\mathcal{A}} \frac{p!}{\prod_{i=1}^n a_i!}\prod_{i=1}^n \E[Y_i^{a_i}]
\le (2p)^p \sum_{a\in \mathcal{A}}\prod_{i=1}^n \sigma_i^{a_i}.
$$

By Claim~\ref{clm:sum} below, we know this expectation is bounded by $p^p (3\sigma)^p$. The second part of the lemma follows immediately from Markov's inequality.
\end{proof}

\begin{claim}\label{clm:sum}
Let $\mathcal{A}_{n,p}$ be the set of vectors $a\in \mathbb{N}^n$ where $a_i = 0$ or $a_i \ge 2$, $\sum_{i=1}^n a_i = p$. For any $n, p \ge 0$ and for any $\sigma_1,...,\sigma_n > 0$, we have
$$
\sum_{a\in \mathcal{A}_{n,p}} \prod_{i=1}^n \sigma_i^{a_i} \le (9\sum_{i=1}^n \sigma_i^2)^{p/2}.
$$
\end{claim}

\begin{proof}
We do induction on $n$. When $n \le 1$ this is clearly correct. Let $F(n,p) = \sum_{a\in \mathcal{A}_{n,p}} \prod_{i=1}^n \sigma_i^{a_i}$, then we have
$$
F(n,p) = F(n-1,p) + \sum_{a = 2}^p F(n-1, p - a) \sigma_n^a.
$$

Suppose the claim is true for all $n < z$, let $\sigma' = \sqrt{\sum_{i=1}^{z-1} \sigma_i^2}$, when $n = z$ we have

\begin{align*}
F(z,p) &= F(z-1,p) + \sum_{a = 2}^p F(n-1, p - a) \sigma_n^a\\
& \le (3\sigma')^p + \sum_{a = 2}^p (3\sigma')^{p-a} \sigma_n^a.
\end{align*}

When $\sigma_n \le 2\sigma'$, we know $\sum_{a = 2}^p (3\sigma')^{p-a} \sigma_n^a \le 3 (3\sigma')^{p-2} \sigma_n^2$, hence by Binomial expansion we have
$$
(9(\sigma')^2 + 9 \sigma_n^2)^{p/2} \ge (3\sigma')^p + (3\sigma')^{p-2} \cdot 9\sigma_n^2 \ge F(z,p).
$$

On the other hand, if $\sigma_n \ge 2\sigma'$, then we know all the terms in the summation $\sum_{a = 2}^p (3\sigma')^{p-a} \sigma_n^a$ and $(3\sigma')^p$ are bounded by $(1.5\sigma_n)^p$, therefore
$$
(9(\sigma')^2 + 9 \sigma_n^2)^{p/2} \ge (3\sigma_n)^p \ge (p-1)(2\sigma_n)^p  \ge F(z,p).
$$

In both cases we prove $F(z,p) \le (9\sum_{i=1}^n \sigma_i^2)^{p/2}$, which finishes the induction.
\end{proof}

We also remark that Lemma~\ref{lem:pwiseconcentration} can be generalized to vectors

\begin{corollary}\label{cor:pwiseconcentration:vector}
Let $X_1,X_2,...,X_n$ be random vectors where $\|X_i\|$ is $\sigma_i$-subexponential. Let $\sigma^2 = \sum_{i=1}^n \sigma_i^2$, $X = \sum_{i=1}^n X_i$. If $X_i$'s are $p$-wise independent, for any even $p$
$$
\E[\|X-\E[X]\|^p] \le (3\sigma)^p\cdot (2p)^p.
$$
In particular, for all $t > 1$,
$$
\Pr[\|X-\E[X]\| \ge 6\sigma p t] \le 1/t^p.
$$
\end{corollary}

\begin{proof}
The proof is exactly the same as the proof of Lemma~\ref{lem:pwiseconcentration}. When $X_i$'s are vectors we get exactly the same terms, except the terms have pair-wise inner-products. However, the inner-products $\inner{X_i,X_j}\le \|X_i\|\|X_j\|$ so we only need to argue about the same inequality for $\|X_i\|$'s.
\end{proof}


%% file: extended_experiments.tex
\section{Extended experiment}
\subsection{Verification of conditions}\label{appdix:verification}

\renewcommand{\thefigure}{A.\arabic{figure}}
\setcounter{figure}{0}

\begin{figure}[H]
    \centering
    \def\mydata{
    1,2,3,4,5,6,7,8,9,10,11,12,13,14,15,16,17,18}
    \foreach \x in \mydata
    {
    \begin{subfigure}[t]{0.3\textwidth}
        \centering
        \includegraphics[width=4.5cm]{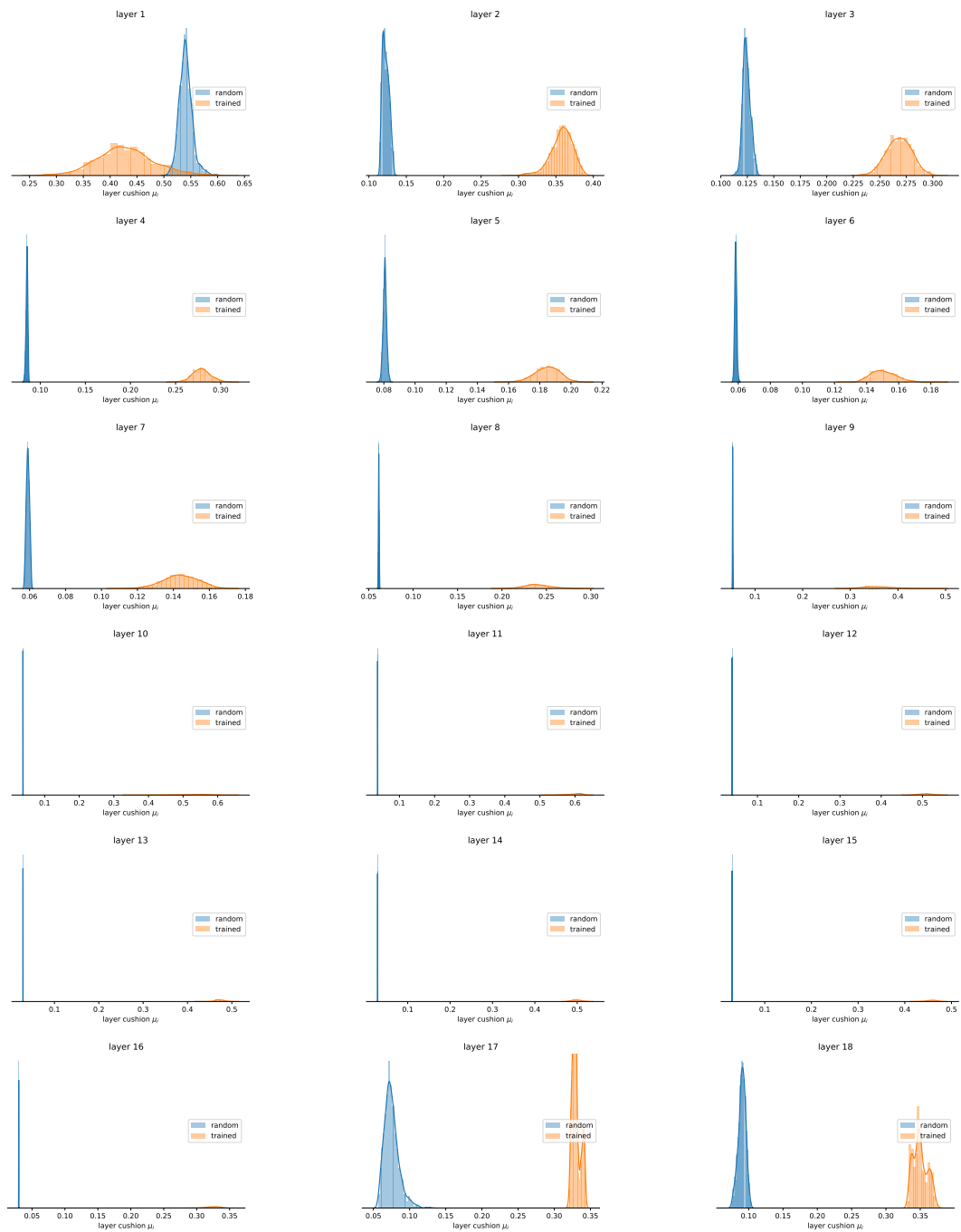}
    \end{subfigure}
    }
    \caption{Verification of layer cushion condition on the VGG-19 net}
\end{figure}

\begin{figure}[H]
    \centering
    \def\mydata{
    1,2,3,4,5,6,7,8,9,10,11,12,13,14,15,16,17,18}
    \foreach \x in \mydata
    {
    \begin{subfigure}[t]{0.3\textwidth}
        \centering
        \includegraphics[width=4.5cm]{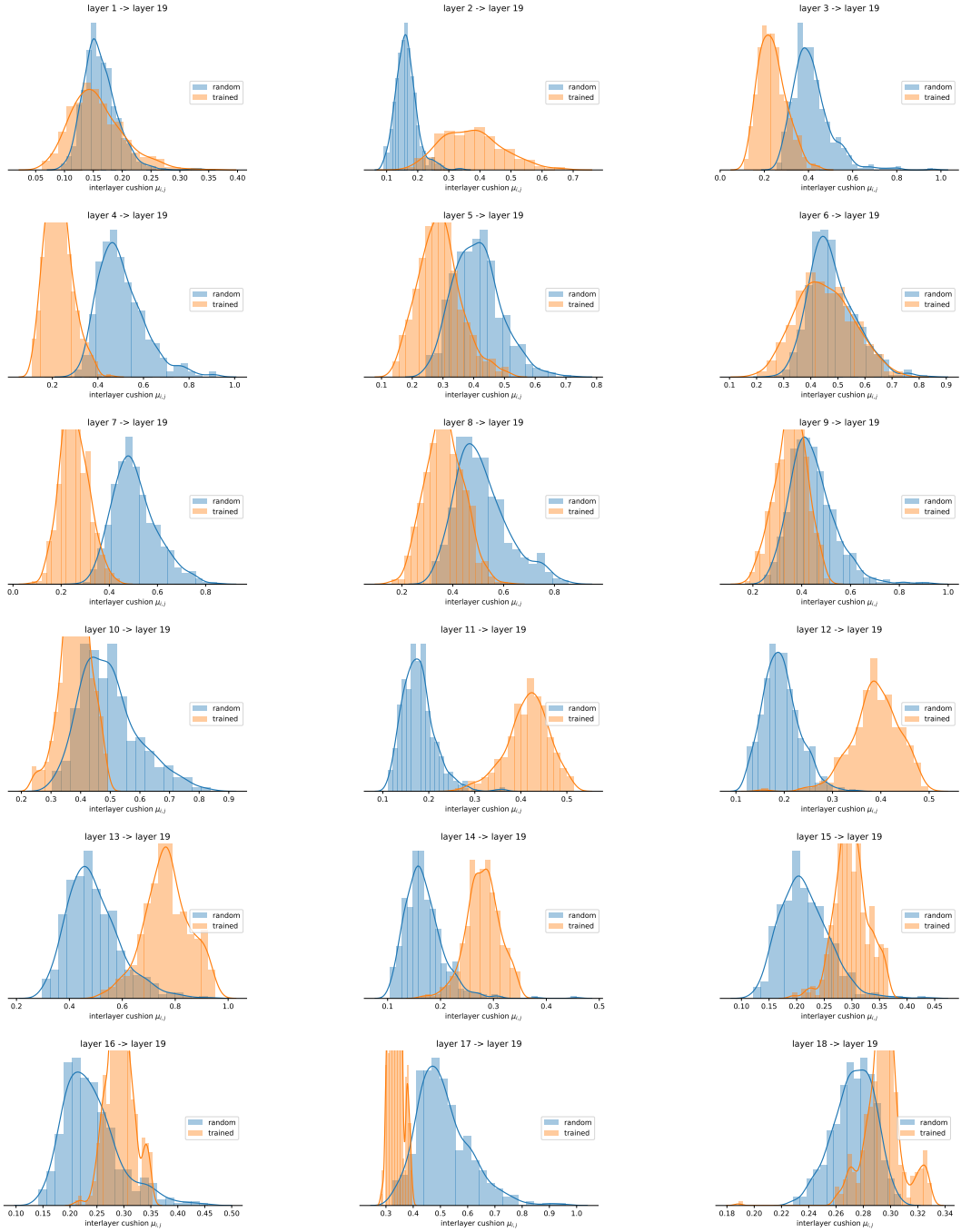}
    \end{subfigure}
    }
    \caption{Verification of interlayer cushion condition on the VGG-19 net}
\end{figure}

\begin{figure}[H]
    \centering
    \def\mydata{
    1,2,3,4,5,6,7,8,9,10,11,12,13,14,15,16,17,18}
    \foreach \x in \mydata
    {
    \begin{subfigure}[t]{0.3\textwidth}
        \centering
        \includegraphics[width=4.5cm]{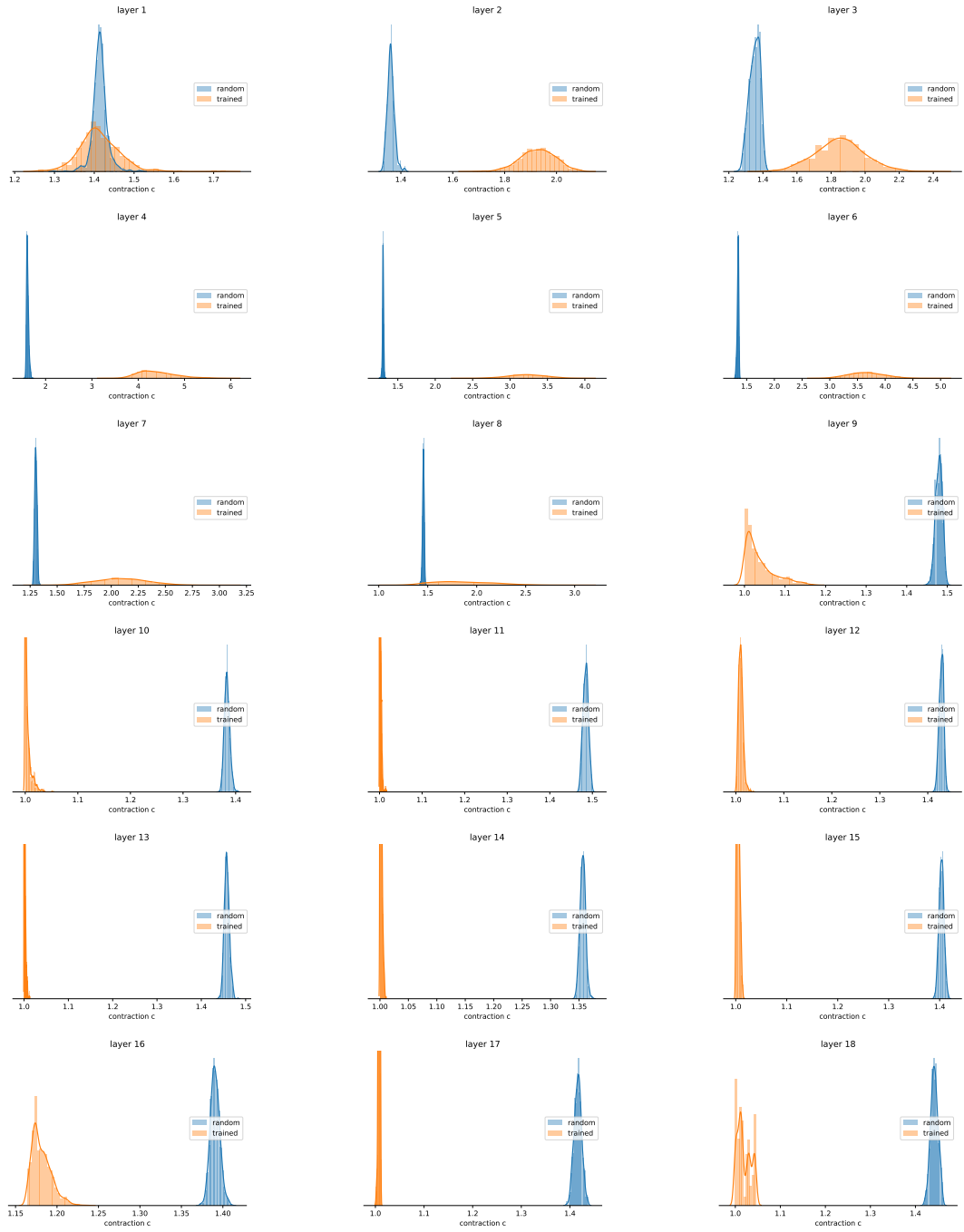}
    \end{subfigure}
    }
    \caption{Verification of activation contraction condition on the VGG-19 net}
\end{figure}

\subsubsection{Verification of interlayer smoothness condition}

\begin{figure}[H]
    \centering
    \def\mydata{
    1,2,3,4,5,6,7,8,9,10,11,12,13,14,15,16,17,18}
    \foreach \x in \mydata
    {
    \begin{subfigure}[t]{0.3\textwidth}
        \centering
        \includegraphics[width=4.5cm]{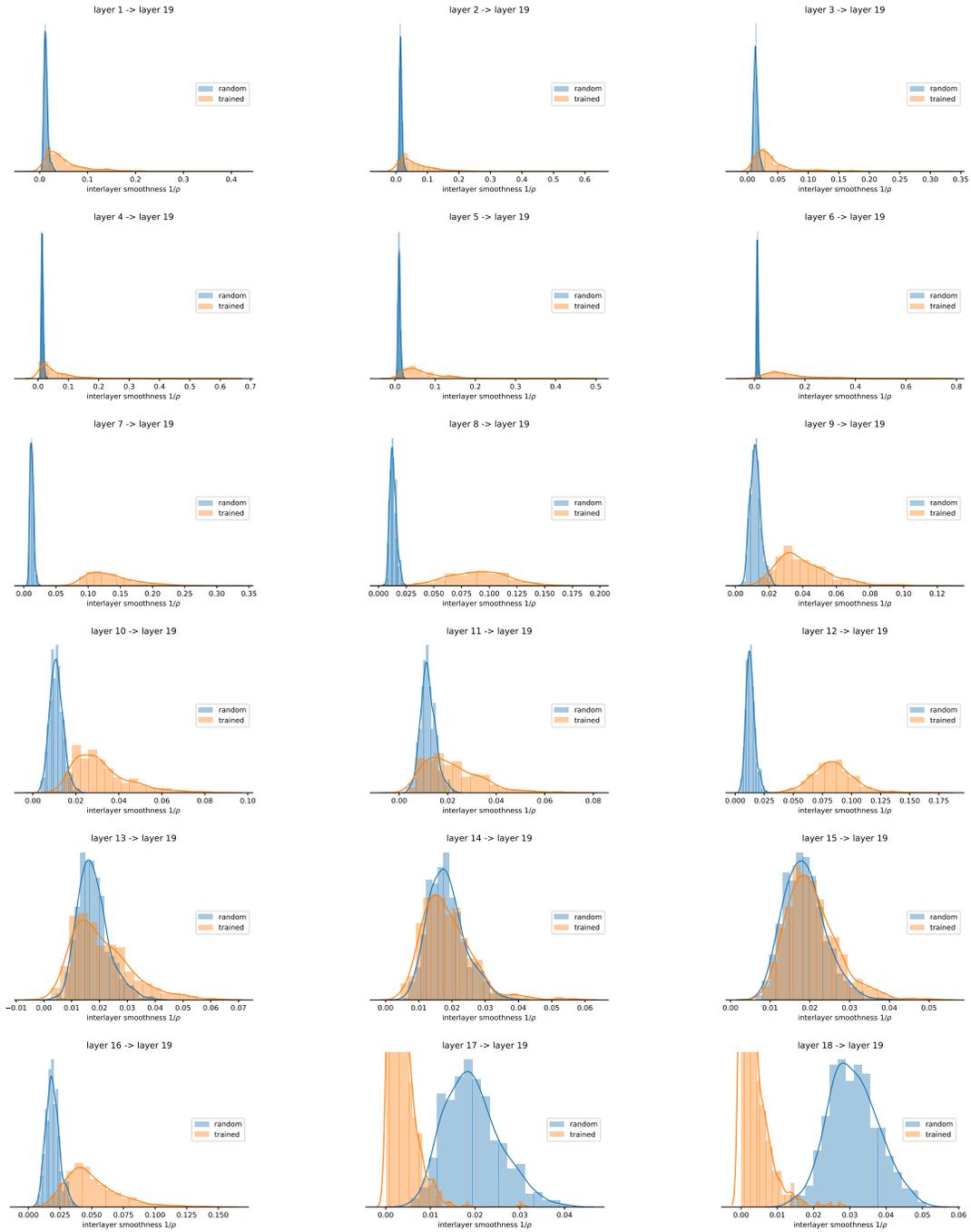}
    \end{subfigure}
    }
    \caption{Verification of interlayer smoothness condition on the VGG-19 net}
\end{figure}

\begin{figure}[H]
    \centering
    \def\mydata{
    1,2,3,4,5,6,7,8,9,10,11,12,13,14,15,16}
    \foreach \x in \mydata
    {
    \begin{subfigure}[t]{0.3\textwidth}
        \centering
        \includegraphics[width=4.5cm]{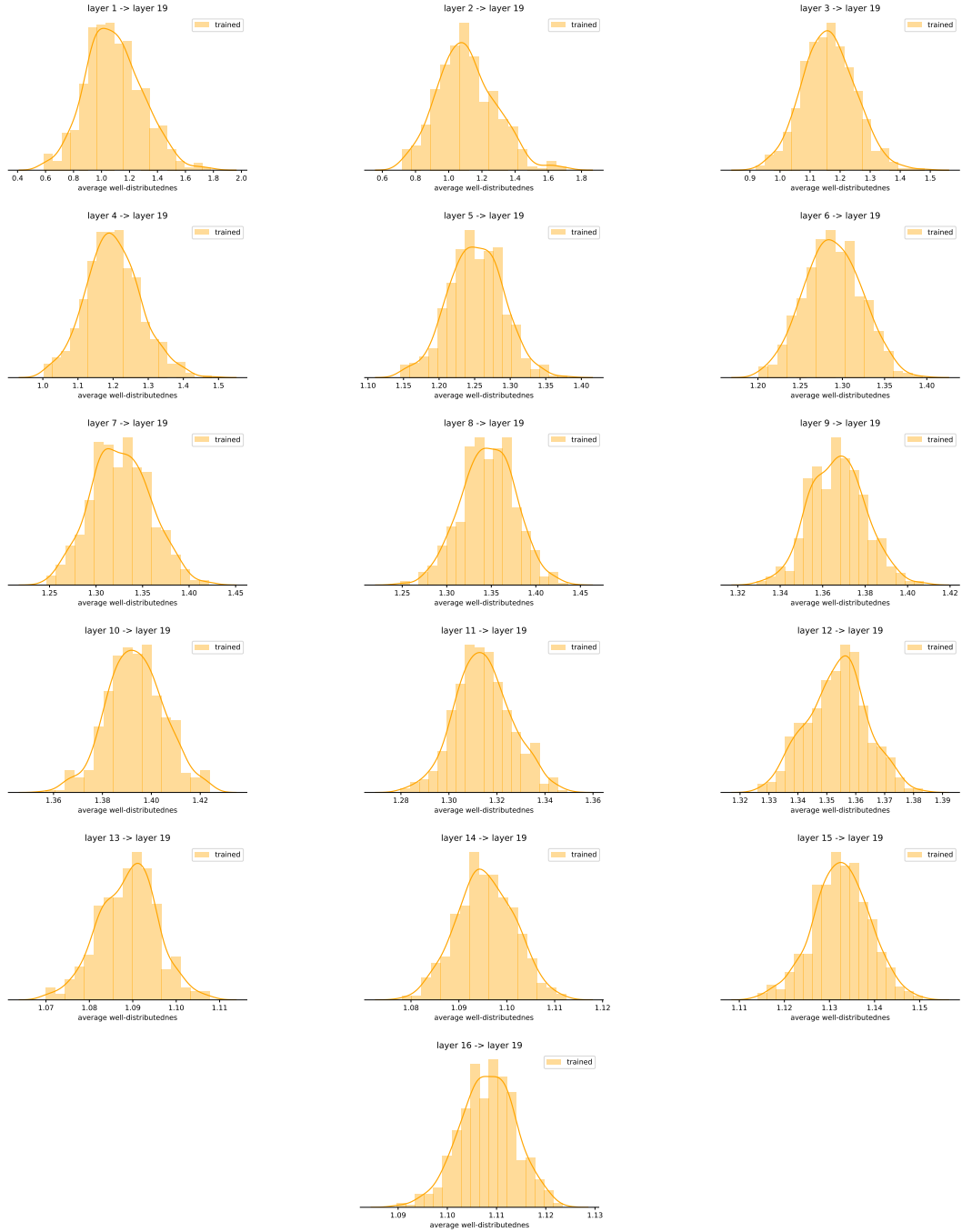}
    \end{subfigure}
    }
    \caption{Verification of well-distributedness of Jacobian condition on convolutional layers of the VGG-19 net. The histograms are generated by estimating the Frobenius norm of the Jacobians of the maps from certain layers to the final layer, restricted on randomly sampled pixels of the input feature maps. Since the well-distributedness parameter $\beta$ is defined to be the largest over all the pixels, $\beta$ should be read off from the upper tails of the histograms. Note for almost all layers, $\beta\approx1$.}
\end{figure}

\subsection{Effect of training on corrupted dataset}\label{appendix:corrupted}
\begin{figure}[H]
    \centering
    \def\mydata{
    1,2,3,4,5,6}
    \foreach \x in \mydata
    {
    \begin{subfigure}[t]{0.3\textwidth}
        \centering
        \includegraphics[width=4.5cm]{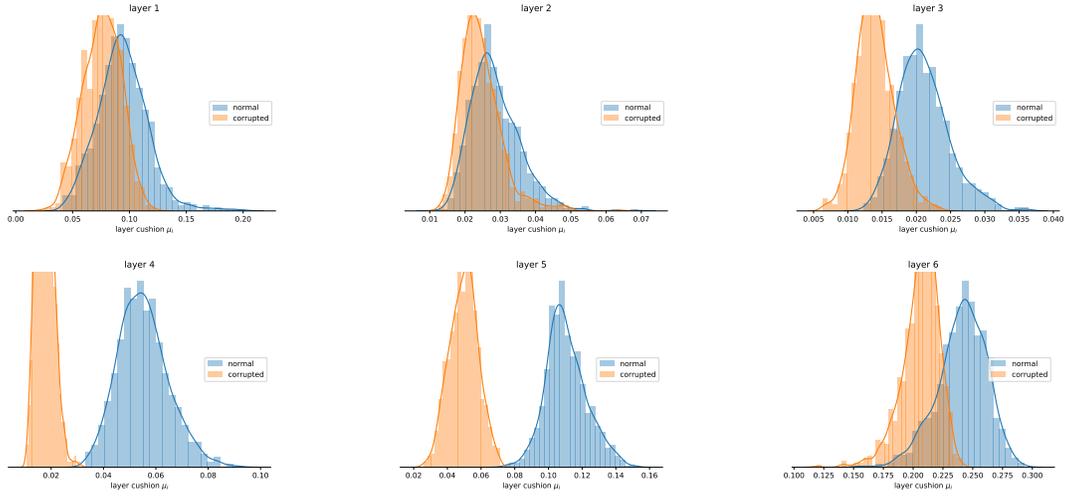}
    \end{subfigure}
    }
    \caption{Distribution of layer cushion of AlexNets trained on normal CIFAR-10 and corrupted CIFAR-10.}
\end{figure}

\begin{figure}[H]
    \centering
    \def\mydata{
    1,2,3,4,5,6}
    \foreach \x in \mydata
    {
    \begin{subfigure}[t]{0.3\textwidth}
        \centering
        \includegraphics[width=4.5cm]{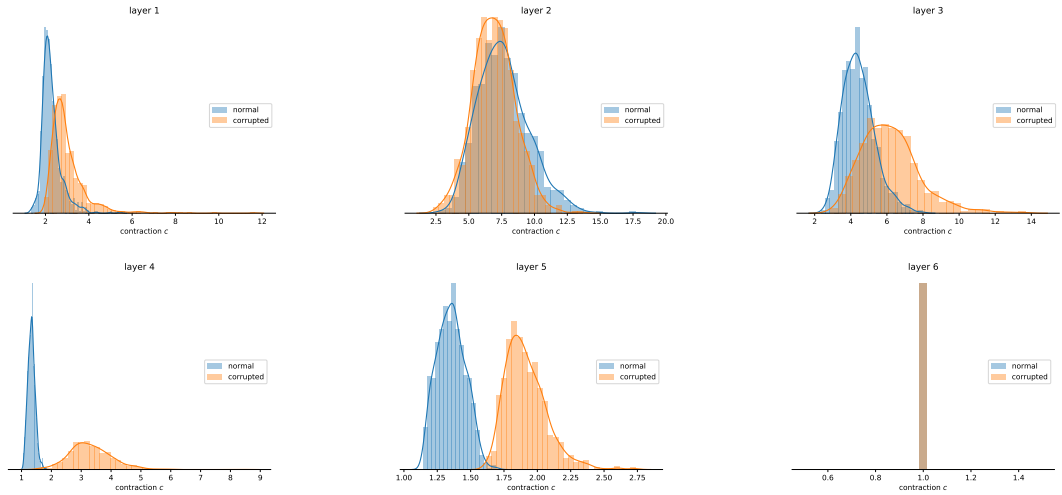}
    \end{subfigure}
    }
    \caption{Distribution of activation contraction of AlexNets trained on normal CIFAR-10 and corrupted CIFAR-10.}
\end{figure}